\PassOptionsToPackage{usenames, dvipsnames}{color}
\PassOptionsToPackage{usenames, dvipsnames}{xcolor}
\documentclass[twoside,11pt]{fairmeta}
\newif\ifdraft
\drafttrue

\usepackage[T1]{fontenc}    %

\usepackage{url}            %
\usepackage{multicol}
\usepackage{hyperref}       %
\usepackage{tabularx}%
\usepackage{siunitx}
\usepackage{graphicx}%
\usepackage{nicefrac}       %
\usepackage{amsmath}
\usepackage{mathtools,amssymb,bm}
\usepackage{amsfonts}
\usepackage{color}
\usepackage{colortbl}
\usepackage{xspace}
\graphicspath{{figures/}}
\usepackage{microtype}      %
\usepackage{geometry}
\usepackage{stmaryrd}
\usepackage{mathrsfs}
\usepackage{algpseudocode}%
\usepackage[inline]{enumitem}
\usepackage{makecell}

\usepackage{tablefootnote}

\usepackage{setspace}
\usepackage{changepage} %
\usepackage{float}

\usepackage{pifont}%

\usepackage{arydshln}
\usepackage[utf8]{inputenc}
\usepackage{hyperref}

\usepackage{algorithm}
\usepackage{algpseudocode}
\usepackage{soul}

\usepackage{siunitx}
\usepackage{placeins}

\newcommand{\llm}{\textsc{LLM}\xspace}
\newcommand{\llms}{\textsc{LLMs}\xspace}
\newcommand{\llama}{\textsc{Llama}\xspace}
\newcommand{\llamatwo}{\textsc{Llama2}\xspace}
\newcommand{\mistral}{\textsc{Mistral}\xspace}
\newcommand{\gemma}{\textsc{Gemma}\xspace}
\newcommand{\bloom}{\textsc{Bloom}\xspace}
\newcommand{\falcon}{\textsc{Falcon}\xspace}
\newcommand{\gpt}{\textsc{GPT}\xspace}
\newcommand{\gemini}{\textsc{Gemini}\xspace}
\newcommand{\claude}{\textsc{Claude}\xspace}
\newcommand{\gemmaIT}{\textsc{Gemma-7B-IT}\xspace}
\newcommand{\mistralIT}{\textsc{Mistral-7B-v0.3-IT}\xspace}
\newcommand{\llamaIT}{\textsc{Llama-3.1-8B-IT}\xspace}
\newcommand{\sllama}{\textsc{smaLlama}\xspace}

\newcommand{\segmenter}{segmenter\xspace}
\newcommand{\segmenters}{segmenters\xspace}

\newcommand{\spacy}{\textsc{SpaCy}\xspace}
\newcommand{\sat}{\textsc{SaT}\xspace}
\newcommand{\jepa}{\textsc{Jepa}\xspace}

\newcommand{\sonar}{\textsc{SONAR}\xspace}
\newcommand{\sonarLangsSpeech}{\textsc{76}\xspace}
\newcommand{\xsim}{\texttt{xsim}\xspace}
\newcommand{\xsimpp}{\texttt{xsim++}\xspace}
\newcommand{\summaryexpansion}{\mbox{\textsc{Summary Expansion}}\xspace}

\newcommand{\lcm}{\textsc{LCM}\xspace}
\newcommand{\lcms}{\textsc{LCMs}\xspace}
\newcommand{\LCM}{\textsc{Large Concept Model}\xspace}
\newcommand{\LCMs}{\textsc{Large Concept Models}\xspace}
\newcommand{\twotower}{\textsc{Two-Tower}\xspace}

\newcommand{\bigtwotower}{\textsc{Two-Tower-7B}\xspace}
\newcommand{\IFTtwotower}{\textsc{Two-Tower-7B-IT}\xspace}

\newcommand{\interleaved}{\textsc{One-Tower}\xspace}
\newcommand{\mselcm}{\textsc{Base-LCM}\xspace}
\newcommand{\qlcm}{\textsc{Quant-LCM}\xspace}
\newcommand{\qlcmd}{\textsc{Quant-LCM-d}\xspace}
\newcommand{\qlcmc}{\textsc{Quant-LCM-c}\xspace}
\newcommand{\planlcm}{\textsc{LPCM}\xspace}

\newcommand{\sonargithub}{\url{https://github.com/facebookresearch/SONAR}}
\newcommand{\meresgithub}{\url{https://github.com/facebookresearch/large_concept_model}}
\newcommand{\stopesgithub}{\url{https://github.com/facebookresearch/stopes}}

\newcommand{\diffx}[2]{\rvx^{#1}_{#2}}

\DeclareMathOperator{\normalize}{normalize}
\DeclareMathOperator{\denormalize}{denormalize}
\DeclareMathOperator{\encode}{encode}
\DeclareMathOperator{\externalencode}{encode_\text{ext}}
\DeclareMathOperator{\decode}{decode}
\DeclareMathOperator{\simil}{score}

\DeclareMathOperator{\fragility}{fragility}
\DeclareMathOperator{\adaln}{AdaLN}
\DeclareMathOperator{\tembed}{embed}
\DeclareMathOperator{\silu}{SiLU}
\DeclareMathOperator{\sigmoid}{sigmoid}
\DeclareMathOperator{\logit}{logit}
\DeclareMathOperator{\tfblock}{Block}
\DeclareMathOperator{\postnet}{PostNet}
\DeclareMathOperator{\prenet}{PreNet}
\DeclareMathOperator{\mse}{MSE}
\newcommand{\sigmainit}{\sigma_\text{init}}
\newcommand{\modeldim}{\rd_\text{model}}
\newcommand{\sonardim}{\rd_\text{\sonar}}

\newcommand{\guidance}{g_\text{scale}}
\newcommand{\rsguidance}{g_\text{rescale}}
\newcommand{\epscaling}{\lambda_\text{eps}}
\newcommand{\ctxenc}{contextualizer\xspace}
\newcommand{\denoiser}{denoiser\xspace}
\newcommand{\ptheta}{\rp_\rvtheta}

\newcommand{\defeq}{\coloneqq}

\DeclareMathOperator{\ltwo}{\ell_2}
\DeclareMathOperator{\ltworound}{\ell_{2-\text{r}}}
\DeclareMathOperator{\paraphrasing}{PAR}
\DeclareMathOperator{\mseacc}{CA}
\DeclareMathOperator{\mutinfo}{MI}
\DeclareMathOperator{\cossim}{CS}
\DeclareMathOperator{\rougel}{R-L}
\DeclareMathOperator{\coherence}{Coherence}

\newcommand{\cnndm}{\textsc{CNN DailyMail}\xspace}
\newcommand{\fineweb}{\textsc{Fineweb-edu}\xspace}
\newcommand{\cfour}{\textsc{C4}\xspace}
\newcommand{\rocstories}{\textsc{ROC-stories}\xspace}
\newcommand{\wikipedia}{\textsc{Wikipedia-en}\xspace}
\newcommand{\gutenberg}{\textsc{Gutenberg}\xspace}
\newcommand{\cosmopedia}{\textsc{Cosmopedia}\xspace}

\newcommand{\flores}{\textsc{Flores}\xspace}
\newcommand{\xsum}{\textsc{XSum}\xspace}
\newcommand{\xlsum}{\textsc{XLSum}\xspace}

\newcommand{\lcfo}{\textsc{LCFO}\xspace}
\newcommand{\lcfofive}{\textsc{LCFO.5\%}\xspace}
\newcommand{\lcfoten}{\textsc{LCFO.10\%}\xspace}
\newcommand{\lcfotwenty}{\textsc{LCFO.20\%}\xspace}

\newcommand{\bleu}{\textsc{BLEU}\xspace}

\newcommand{\autobleu}{\textsc{AutoBLEU}\xspace}

\newcommand{\rougellong}{\textsc{Rouge-L}\xspace}
\newcommand{\ovlthree}{\textsc{OVL-3}\xspace}
\newcommand{\repfour}{\textsc{REP-4}\xspace}
\newcommand{\cola}{\textsc{CoLA}\xspace}
\newcommand{\shfour}{\textsc{SH-4}\xspace}
\newcommand{\shfive}{\textsc{SH-5}\xspace}

\newcolumntype{H}{>{\setbox0=\hbox\bgroup}c<{\egroup}@{}}
\newcolumntype{P}[1]{>{\centering\arraybackslash}p{#1}}
\newcommand{\MC}{\multicolumn}
\newcommand{\MR}{\multirow}

\newcommand{\fairseq}{\textsc{Fairseq2}\xspace} %
\newcommand{\ie}{\textit{i.e.,}\xspace}
\newcommand{\eg}{\textit{e.g.,}\xspace}

\definecolor{alizarin}{rgb}{0.82, 0.1, 0.26}
\definecolor{asparagus}{rgb}{0.53, 0.66, 0.42}
\definecolor{cerulean}{rgb}{0.0, 0.48, 0.65}

\newcommand{\cmark}{\color{ForestGreen}{\ding{51}}}%
\newcommand{\xmark}{\color{BrickRed}{\ding{55}}}%

\newcommand{\ds}{\displaystyle}

\newcommand{\loss}{\mathcal{L}}

\newcommand{\gaussian}{\mathcal{N}}
\newcommand{\uniform}{\mathcal{U}}

\def\rd{{\textnormal{d}}}

\def\rF{{\textnormal{F}}}
\def\rg{{\textnormal{g}}}

\def\rN{{\textnormal{N}}}

\def\rp{{\textnormal{p}}}
\def\rq{{\textnormal{q}}}

\def\rs{{\textnormal{s}}}
\def\rS{{\textnormal{S}}}

\def\rT{{\textnormal{T}}}

\def\rvalpha{{\bm{\alpha}}}
\def\rvbeta{{\bm{\beta}}}
\def\rvgamma{{\bm{\gamma}}}
\def\rvepsilon{{\bm{\epsilon}}}
\def\rvsigma{{\bm{\sigma}}}
\def\rvtheta{{\bm{\theta}}}
\def\rvmu{{\bm{\mu}}}

\def\rvb{{\mathbf{b}}}
\def\rvc{{\mathbf{c}}}

\def\rvr{{\mathbf{r}}}

\def\rvx{{\mathbf{x}}}
\def\rvy{{\mathbf{y}}}

\def\rvzero{{\mathbf{0}}}

\def\rmI{{\mathbf{I}}}

\def\rmW{{\mathbf{W}}}

\def\rmSigma{{\mathbf{\Sigma}}}

\usepackage{colortbl}
\colorlet{tableheadcolor}{gray}
\colorlet{tablerowcolor}{gray!10}

\author[]{LCM team}
\author[*]{Loïc Barrault} 
\author[*]{Paul-Ambroise Duquenne} 
\author[*]{Maha Elbayad} 
\author[*]{Artyom Kozhevnikov}
\author[\dagger]{Belen Alastruey}
\author[\dagger]{Pierre Andrews}
\author[\dagger]{Mariano Coria}
\author[+\dagger]{Guillaume Couairon}
\author[\dagger]{Marta R. Costa-jussà}
\author[\dagger]{David Dale}
\author[\dagger]{Hady Elsahar}
\author[\dagger]{Kevin Heffernan}
\author[\dagger]{João Maria Janeiro}
\author[\dagger]{Tuan Tran}
\author[\dagger]{Christophe Ropers}
\author[\dagger]{Eduardo S\'anchez}
\author[\dagger]{Robin San Roman}
\author[\ddagger]{Alexandre Mourachko} 
\author[\ddagger]{Safiyyah Saleem} 
\author[\ddagger]{Holger Schwenk} 

\affiliation[]{FAIR at Meta}
\contribution[*]{Core contributors, alphabetical order}
\contribution[\dagger]{Contributors to data preparation, LCM extensions and evaluation, alphabetical order}
\contribution[\ddagger]{Research and project management, alphabetical order}
\contribution[+]{Initial work while at FAIR at Meta, new affiliation: INRIA, France}

\date{December 12, 2024}
\correspondence{Holger Schwenk at \email{schwenk@meta.com}}

\title{{\LCMs:}\\ \Large
Language Modeling in a Sentence Representation Space
}

\abstract{
\small
\llms have revolutionized the field of artificial intelligence and have emerged as the de-facto tool for many tasks. The current established technology of \llms is to process input and generate output at the token level.
This is in sharp contrast to humans who operate at multiple levels of abstraction, well beyond single words, to analyze information and to generate creative content.
In this paper, we present an attempt at an architecture which operates on an explicit higher-level semantic representation, which we name a \textit{``concept''}. Concepts are language- and modality-agnostic and represent a higher level idea or action in a flow. Hence, we build a \textit{``\LCM''}.
In this study, as proof of feasibility, we assume that a concept corresponds to a sentence, and use an existing sentence embedding space, \sonar, which supports up to 200 languages in both text and speech modalities.

The \LCM is trained to perform autoregressive sentence prediction in an embedding space.
We explore multiple approaches, namely MSE regression, 
variants of diffusion-based generation, 
and models operating in a quantized \sonar space.
These explorations are performed using 1.6B parameter models and training data in the order of 1.3T tokens.
We then scale one architecture to a model size of 7B parameters and training data of about 2.7T tokens.
We perform an experimental evaluation on several generative tasks, namely summarization and a new task of summary expansion.
Finally, we show that our model exhibits impressive zero-shot generalization performance to many languages, outperforming existing \llms of the same size.
The training code of our models is freely available.\footnote{\meresgithub}}

\begin{document}

\maketitle

\section{Introduction}
\label{sec:introduction}
Large Language models (\llms) are dominating current research in natural language processing, and with their recent extension to more modalities, namely images, video and speech, they seem to be considered as the de-facto technique to follow to approach human intelligence. \llms achieve indeed impressive performance on a large variety of tasks, such as  providing detailed answers for general knowledge questions, helping in performing long document analysis, or drafting different types of messages, and writing or debugging code.
Building an \llm from scratch requires access to enormous computational resources to process ever larger amounts of data and train models, the size of which now exceeds four hundred billion parameters.
Knowledge acquisition in \llms is heavily data-driven and extending them to more languages or modalities usually requires injecting additional (synthetic) data to cover them. 

The landscape of available \llms can be structured into open models such as \llama\citep{llama3.arxiv.2024}, \mistral \citep{mixtral.arxiv.2024}, \bloom \citep{bloom.arxiv.2023} or \falcon \citep{falcon.arxiv.2023}, on the one hand,
and closed models such as \gemini \citep{gemini15.arxiv.2024}, \gpt \citep{gpt4.arxiv.2024} or \claude \citep{claude3.anthropic.2024}, on the other.
It is striking that all these models are based on the same underlying architecture: a transformer-based, decoder-only language model, which is pretrained to predict the next token, given a long context of preceding tokens.
Despite the undeniable success of \llms and continued progress, all current \llms miss a crucial characteristic of human intelligence: explicit reasoning and planning at multiple levels of abstraction.
The human brain does not operate at the word level only.
We usually have a top-down process to solve a complex task or compose a long document: we first plan at a higher level the overall structure, and then step-by-step, add details at lower levels of abstraction.
One may argue that \llms are implicitly learning a hierarchical representation, but we stipulate that models with an explicit hierarchical architecture are better suited to create coherent long-form output.

Imagine a researcher giving a fifteen-minute talk. In such a situation, researchers do not usually prepare detailed speeches by writing out every single word they will pronounce. Instead, they outline a flow of higher-level ideas they want to communicate. Should they give the same talk multiple times, the actual words being spoken may differ, the talk could even be given in different languages, but the flow of higher-level abstract ideas will remain the same.
Similarly, when writing a research paper or essay on a specific topic, humans usually start by preparing an outline that structures the whole document into sections, which they then refine iteratively.
Humans also detect and remember dependencies between the different parts of a longer document at an abstract level. If we expand on our previous research writing example, keeping track of dependencies means that we need to provide results for each of the experiment mentioned in the introduction. Finally, when processing and analyzing information, humans rarely consider every single word in a large document. Instead, we use a hierarchical approach: we remember which part of a long document we should search to find a specific piece of information.

To the best of our knowledge, this explicit hierarchical structure of information processing and generation, at an abstract level, independent of any instantiation in a particular language or modality, cannot be found in any of the current \llms.
\begin{figure}
     \begin{tabular}[c]{c}
         \includegraphics[width=0.5\textwidth]{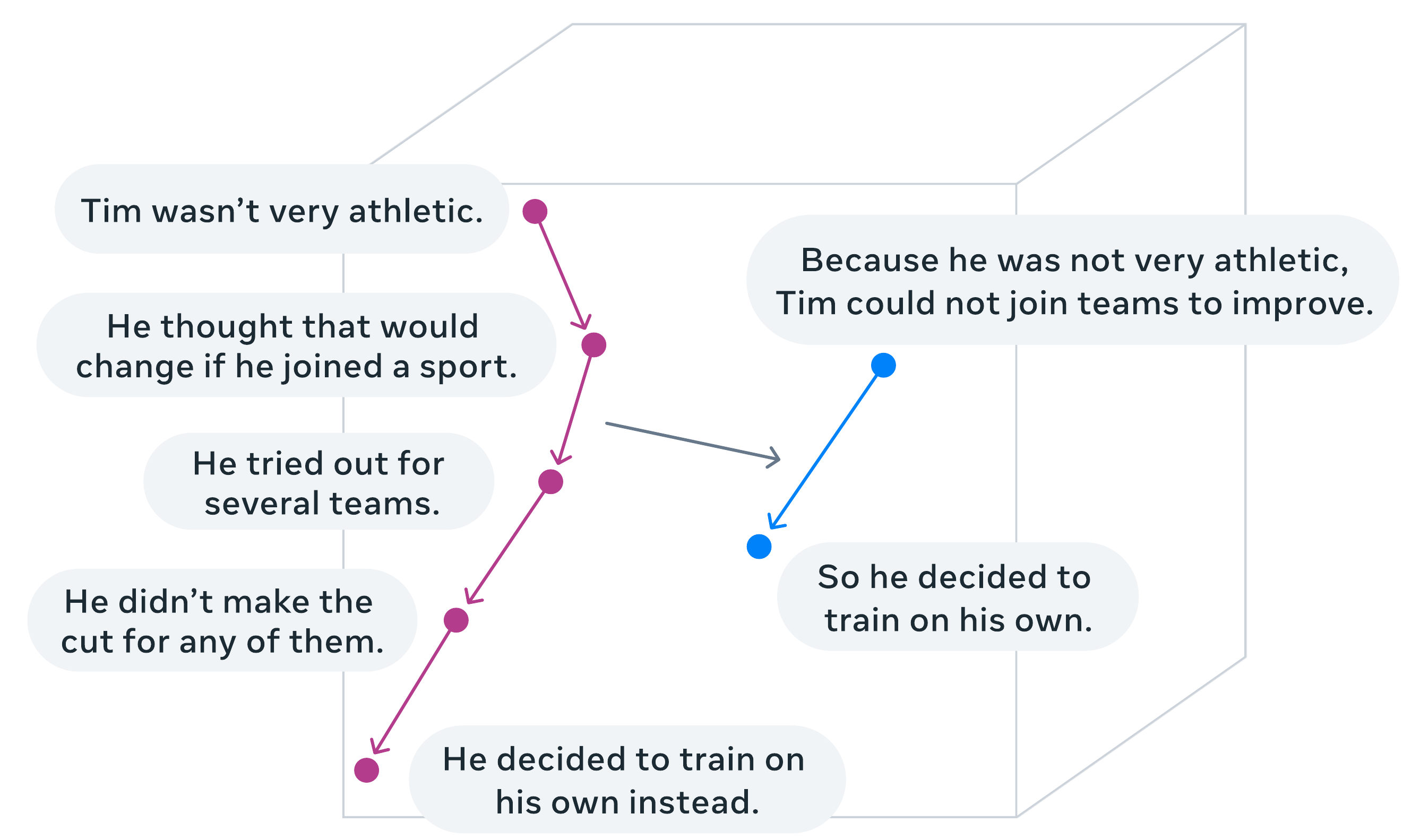}
     \end{tabular}
     \hfill
     \begin{tabular}[c]{c}
         \includegraphics[width=0.45\textwidth]{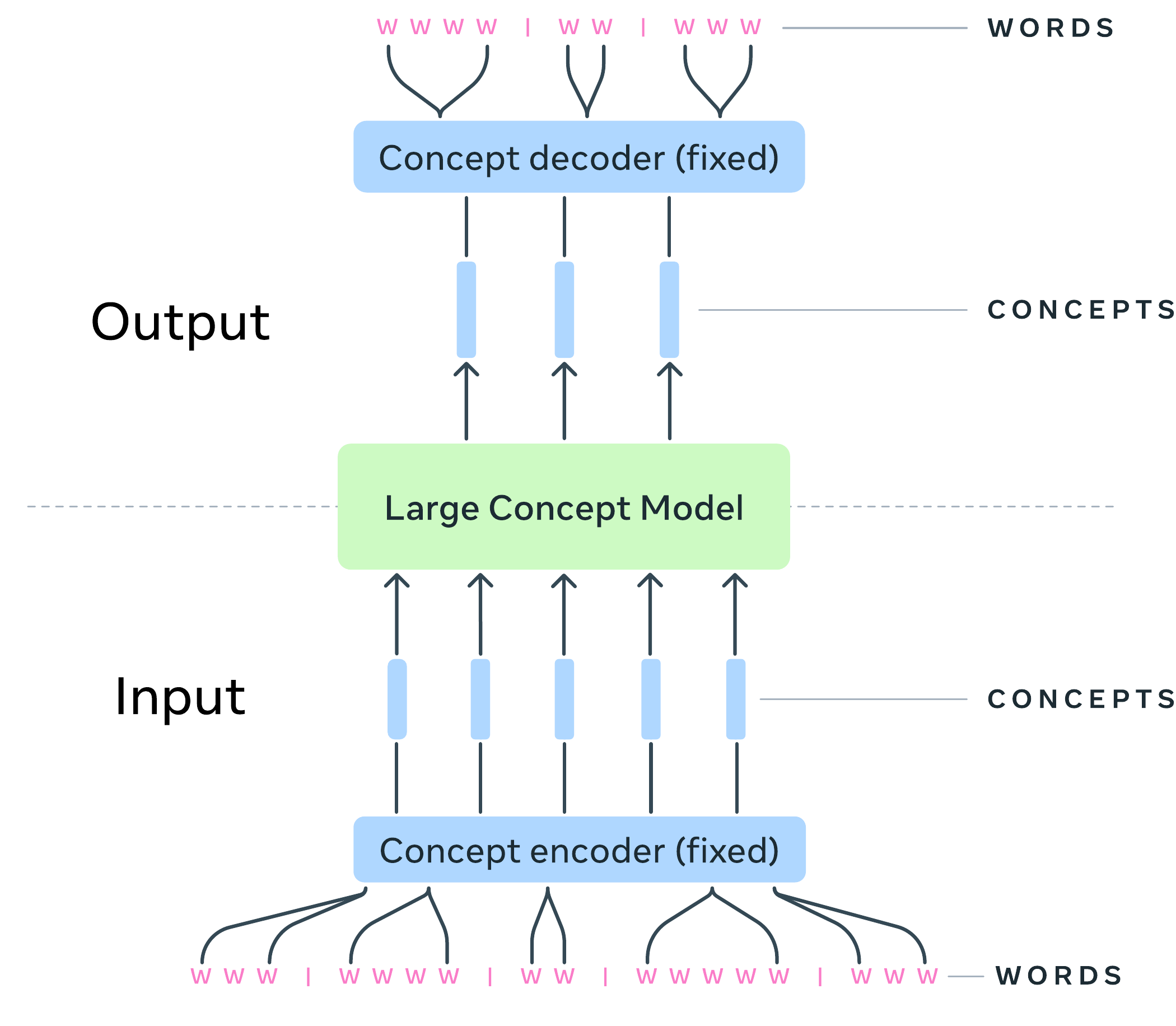}
     \end{tabular}
    \caption{Left: visualization of reasoning in an embedding space of concepts (task of summarization). \\ Right: fundamental architecture of an \LCM (\lcm).\\$\star$: concept encoder and decoder are frozen.}
    \label{fig:intro:idea}
\end{figure}
In this work, we present a new approach which moves away from processing at the token level and closer to (hierarchical) reasoning in an abstract embedding space. This abstract embedding space is designed to be independent of the language or modality in which the content is expressed; in other words, we aim to model the underlying reasoning process at a purely semantic level, not its instantiation in a specific language. 
In order to verify our approach, we limit our study to two levels of abstraction: subword tokens and \textit{concepts}. We define a \textit{concept} as an abstract atomic idea.
In practice, a concept would often correspond to a sentence in a text document, or an equivalent speech utterance. We posit that a sentence is an appropriate unit to achieve language independence, in opposition to single words.
This is in sharp contrast to current \llms techniques which are heavily English centric and token based.

Our fundamental idea could be based on any fixed-size sentence embedding space for which an encoder and decoder are available. In particular, we could aim to train a new embedding space specifically optimized to our reasoning architecture. In this work, we chose an existing and freely available sentence embedding, named \sonar \citep{Duquenne:2023:sonar_arxiv}. \sonar supports text input and output in 200 languages, speech input in \sonarLangsSpeech languages, and speech output in English. We discuss the constraints and impact of this choice in \Cref{sec:archi:sonar}, and share some ideas on alternative embedding spaces in \Cref{sec:limits}.

\Cref{fig:intro:idea}-left visualizes reasoning in an embedding space with the example of a summarization task, which is materialized by a function on the embedding space, mapping five concept representations into two.
\Cref{fig:intro:idea}-right summarizes the overall architecture and processing flow. The input is first segmented into sentences, and each one is encoded with \sonar to achieve a sequence of concepts, \ie sentence embeddings.
This sequence of concepts is then processed by a \textbf{\LCM (\lcm)} to generate at the output a new sequence of concepts. Finally, the generated concepts are decoded by \sonar into a sequence of subwords. The encoder and decoder are fixed and are not trained. It is important to highlight that the unchanged sequence of concepts at the output of the \lcm  can be decoded into other languages or modalities without performing again the whole reasoning process. In the same spirit, a particular reasoning operation such as summarization can be performed in a zero-shot setting on input in any language or modality, since it solely operates on concepts. To summarize, the \lcm neither has information on the input language or modality nor generates output in a particular language or modality.
We explore multiple architectures to train the \lcm, in particular several variants of diffusion.
Finally, we envision an additional level of abstraction beyond concepts which could correspond to a short description of a paragraph or small section. In \Cref{sec:planlcm} we report initial ideas on how conditioning and predicting such higher-level representations can improve consistency of output generated by an \lcm.

To some extent, the \lcm architecture resembles the \jepa approach \citep{jepa:openreview:2022} that also aims to predict the representation of the next observation in an embedding space. However, unlike \jepa that places more emphasis on learning a representation space in a self-supervised way, the \lcm focuses on accurate prediction in the existing embedding space.

\newpage %
The mains characteristics of our generic \LCM approach are as follows:
\begin{itemize}[style=unboxed,leftmargin=*]
    \item \textbf{Reasoning at an abstract language- and modality-agnostic level beyond tokens:}
        \begin{itemize}[style=unboxed,leftmargin=*]
             \item We model the underlying reasoning process, not its instantiation in a particular language.
             \item The \lcm can be trained, i.e. acquire knowledge, on all languages and modalities at once, promising scalability in an unbiased way.
        \end{itemize}
    \item \textbf{Explicit hierarchical structure:}
        \begin{itemize}[style=unboxed,leftmargin=*]
            \item Better readability of long-form output by a human.
            \item Facilitates local interactive edits by a user.
        \end{itemize}
    \item \textbf{Handling of long context and long-form output:}
        \begin{itemize}[style=unboxed,leftmargin=*]
            \item The complexity of a vanilla transformer model increases quadratically with the sequence length. This makes handling of large context windows challenging and several techniques have been developed to alleviate this problem, \eg sparse attention \citep{child2019generating} or LSH attention \citep{kitaev2020reformer}.
            Our \lcm operates on sequences which are at least an order of magnitude shorter.\footnote{We assume an average sentence length of 10--20 tokens.}
        \end{itemize}
    \item \textbf{Unparalleled zero-shot generalization:}
        \begin{itemize}[style=unboxed,leftmargin=*]
            \item Independently of the language or modality the \lcm is pre-trained and fine-tuned on, it can be applied to any language and modality supported by the \sonar encoders, without the need of additional data or fine-tuning. We report results for multiple languages in the text modality.
        \end{itemize}
    \item \textbf{Modularity and extensibility:}
        \begin{itemize}[style=unboxed,leftmargin=*]
            \item Unlike multimodal \llms that can suffer from modality competition~\citep{aghajanyan2023scaling,chameleon:2024:arxiv}, concept encoders and decoders can be independently developed and optimized without any competition or interference. 
            \item New languages or modalities can be easily added for an existing system.
        \end{itemize}
\end{itemize}

The goal of this paper is to provide a proof of concept of this high-level vision of an alternative architecture to current best practice in language modeling.
In the next section we present the main design principles of our models and discuss several variants to build and train a \LCM.
We discuss several designs to implement diffusion approaches with concept embeddings and carefully study noise scheduling. This section is completed by a compute complexity comparison with token-based \llms.
\Cref{sec:bigmodel} is dedicated to the analysis of a larger 7B parameter model. We discuss challenges when instruction fine-tuning this model on multiple generative tasks, and provide a comparison with existing \llms of comparable size.
The paper concludes with a discussion of related work, the current limitations and perspectives of our approach.

To foster research in this area, we make our \lcm training code\footnote{\meresgithub} as well as SONAR encoders and decoders\footnote{\sonargithub} for up to 200 languages and multiple modalities freely available.

\newpage
\section{Main Design Principles}

In this section, we outline the main design principles of the \lcm. 
We first describe the \sonar embedding space with its encoders and decoders.
Then, we discuss details of data preparation, namely sentence segmentation \ie how we split long documents into sentences. 
And finally, we describe in details the different versions of \lcms introduced in this work.

\subsection{The \sonar embedding space}
\label{sec:archi:sonar}

The motivation of this work is to perform reasoning at a higher conceptual level than tokens. This requires an embedding space which is highly semantic. We chose \sonar \citep{{Duquenne:2023:sonar_arxiv}} since it achieves best performance on several semantic similarity metrics like \xsim or \xsimpp \citep{xsimpp:arxiv:2023}, and it was successfully used in large-scale bitext mining for translation~\citep{SeamlessM4TArXiv}.

\begin{figure}[!b]
    \centering
    \includegraphics[width=0.8\linewidth]{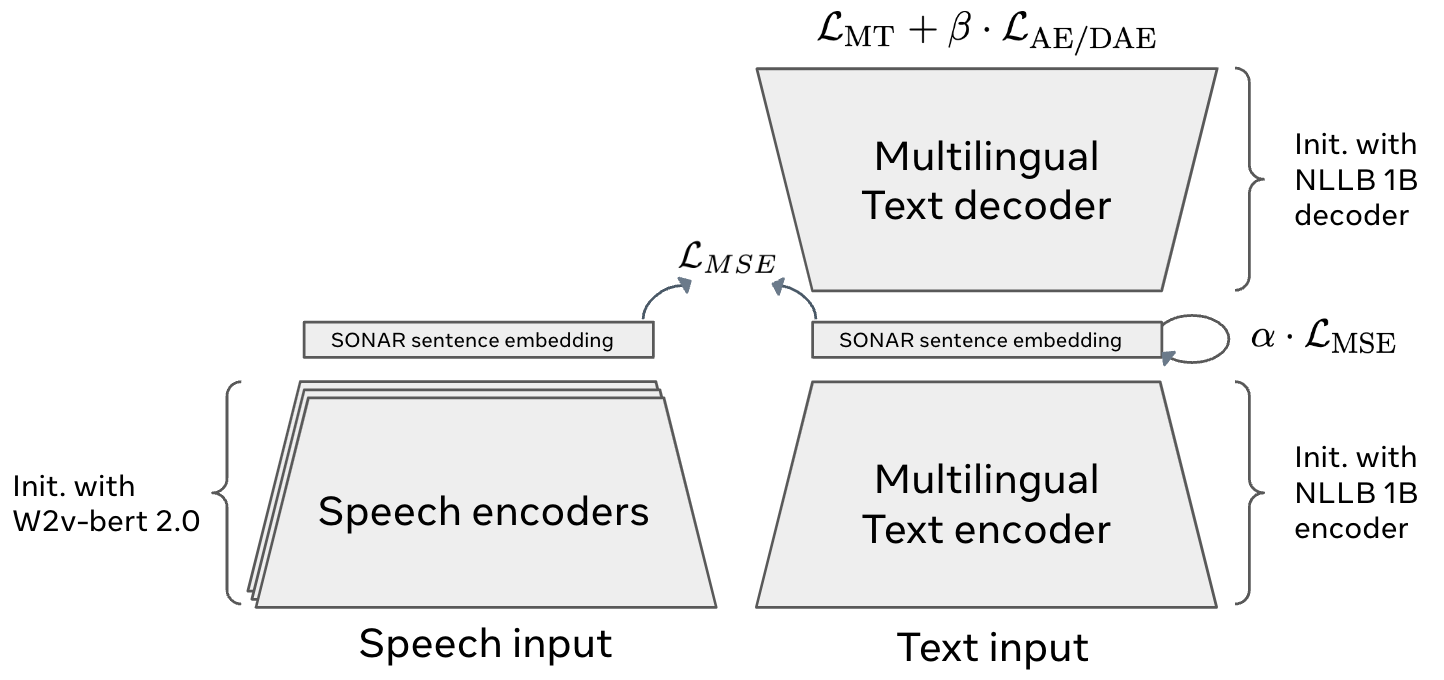}
    \caption{Encoder/decoder bottleneck architecture to train the \sonar text embeddings (right part of figure). Teacher-student approach to extend \sonar to the speech modality (left part).} 
    \label{fig:archi:sonar}
\end{figure}
The \sonar text embedding space was trained as an encoder/decoder architecture, with a fixed-size bottleneck instead of cross-attention (see \Cref{fig:archi:sonar}). The criterion combines a machine translation objective for 200 languages into and out of English, denoising auto-encoding and an explicit MSE loss at the embedding bottleneck layer.
Once the text embedding space was trained, a teacher-student approach was applied to extend the \sonar space to the speech modality. 
More details on the architecture and training procedure can be found in \citet{Duquenne:2023:sonar_arxiv}, and detailed speech recognition and translation results in the appendix of \citet{seamlessv2:arxiv:2023}.

\begin{table}[!ht]
    \centering
    \begin{tabular}{r|*{8}{c}}
        \toprule
         & \MC{2}{c}{Text} & \MC{2}{c}{Speech} & \MC{2}{c}{Image} & \MC{2}{c}{Video} \\
        Model & Input & Output & Input & Output & Input & Output & Input & Output \\
        \midrule
        \gemini & 47 & 47 & 62 & \cmark & \cmark & \cmark & \cmark & \xmark \\
        \gpt & 85 & 85 & \cmark & \cmark & \cmark & \cmark & ? & \xmark \\
        \claude
         & 37 & 37 & \cmark & \cmark & \cmark & \cmark & \xmark & \xmark \\
        \bloom
         & 46 & 46 & \xmark & \xmark & \cmark & \cmark & \xmark & \xmark \\
        \llama3-400B & 8 & 8 & 34 & \xmark & \cmark & \cmark & \xmark & \xmark\\
        \midrule
        \lcm-\sonar & 200 & 200 & \sonarLangsSpeech & 1 & \xmark & \xmark & (ASL) & \xmark \\
        \bottomrule
    \end{tabular}
   \caption{Comparison of language and modality coverage for several \llms and our \lcm operating on the \sonar embedding space. \sonar has an experimental support for American Sign Language (ASL) which is not used in this paper.
   }
    \label{fig:archi:langs}
\end{table}
Our \lcm operates directly on \sonar concepts embeddings, hence, it can perform reasoning on all supported languages and modalities. \Cref{fig:archi:langs} compares the language coverage of several other \llms. The \lcm supports substantially more languages than other models, in particular many low-resource languages. In addition to the text modality, \sonar supports \sonarLangsSpeech languages for speech input and speech output in English. We have also developed an experimental encoder for American Sign language (ASL).
All these encoders and decoders are freely available.\footnote{\url{https://github.com/facebookresearch/SONAR}}
Exact listings of the supported languages can be found in the \sonar GitHub repository.

\newpage
\FloatBarrier
\subsection{Data preparation}
\label{sec:data}

To train and evaluate the \lcm, we need to convert raw text datasets into a sequence of \sonar embeddings, each one corresponding to a sentence.
Dealing with large text corpora presents several practical limitations.
First, the precise segmentation of a text into sentences can be challenging due to the presence of errors, specific formatting issues or any other sources of noise.
This requires us to apply robust automatic text segmentation techniques.
Second, some sentences (even well formed) can be very long and complex, which might negatively impact the quality of the encoded \sonar embeddings.
This is particularly prevalent for texts in the  scientific domain.
In the following, we discuss strategies for sentence segmentation and
how they affect the \sonar encoding.

\vspace{-3mm} %
\paragraph{Sentence segmentation analysis}
\label{sec:data-analysis}

\label{sec:archi:sentence_seg}

We have identified two potential sentence segmentation techniques; as we are exploring multilingual data, we focus on sentence \segmenters with a large language coverage:
\begin{enumerate}
    \item SpaCy \segmenter (\spacy)~\citep{Honnibal_spaCy_Industrial-strength_Natural_2020} is a well established multilingual NLP toolkit that provides a rule-based approach to sentence segmentation. \spacy is thoroughly tested for high-resource languages.
    \item Segment any Text (\sat)~\citep{minixhofer-etal-2023-wheres,frohmann-etal-2024-segment} offers a suite of models and adapters that predict sentence boundaries at the token level.
    \sat is designed to be resilient to perturbations, particularly avoiding the over-reliance on punctuation and capitalization. This is valuable in domains where these conventional markers are often missing. The quality of \sat's segmentation is however dependent on the choice of an ``appropriate'' split probability threshold.
\end{enumerate}

We additionally customize both methods by incorporating a maximum sentence length cap in characters. We refer to these extensions by \spacy Capped and \sat Capped.
Long sentences are broken down into smaller, logically coherent fragments using a rule-based approach based on punctuation marks for \spacy. For \sat, we leverage the provided splitting probability estimates to identify the next best potential split.

To measure the efficacy of a given \segmenter,
we evaluate the quality of the reconstructed sentences with \autobleu. 
It is defined as a \bleu score~\citep{papineni2002bleu} comparing 
the decoded text from a \sonar vector after encoding a segment,
to the the reference segment.  
A good segmentation will yield segments that can be encoded and then decoded without loss of signal, and thus score a higher \autobleu.

For this analysis, we sample 10k documents from our pretraining datasets, representing approximately 500k sentences. 
The documents are processed with each \segmenter, the sentences are encoded then decoded and the \autobleu score is calculated.
We stratified the results based on the lengths of the original sentences.

\begin{figure}[!htb]
    \centering
    \includegraphics[width=.855\linewidth]{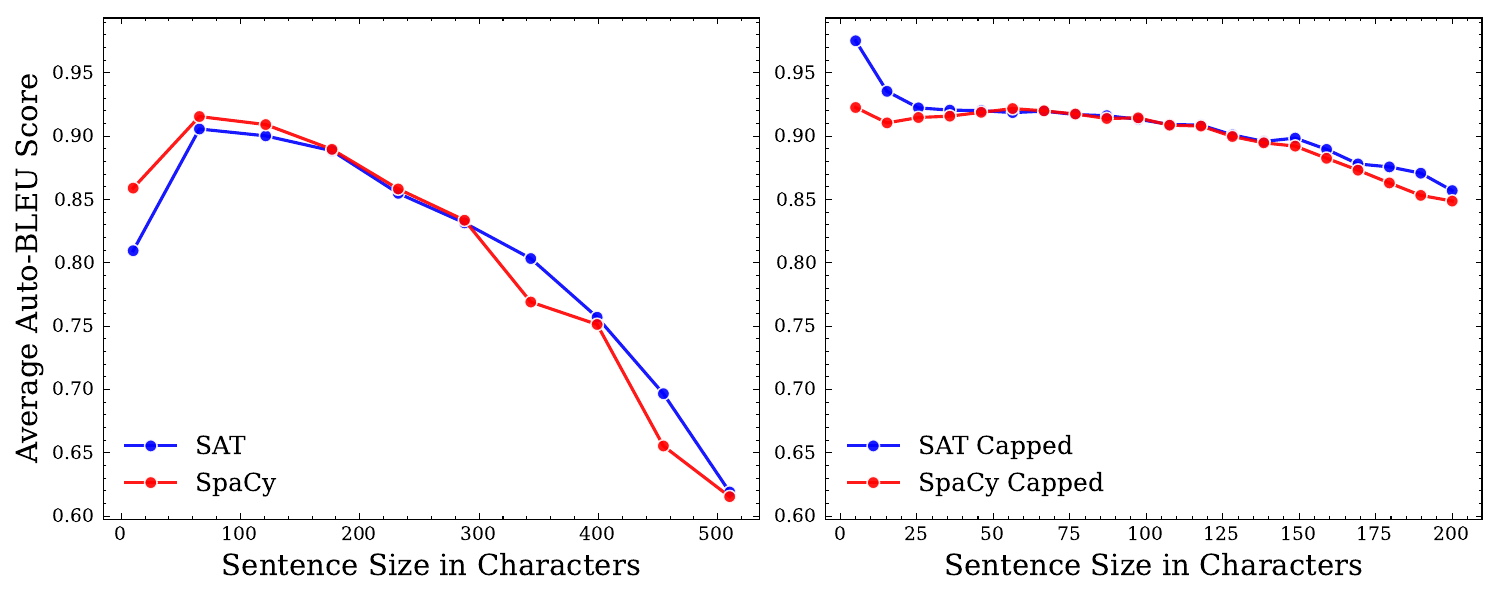}
     
    \caption{\textbf{Segmenters quality.} Average Auto-BLEU scores for different sentence segmentation methods depending on sentence length, for both out of the box (left) and capped implementations (right).
    }
    \label{fig:sent_segmentation}
\end{figure}

As illustrated in \Cref{fig:sent_segmentation} and with a capping at 200 characters, the \sat Capped method demonstrates a slight but consistent advantage over \spacy Capped. 
Both out-of-the-box \segmenters, however, exhibit significant under-performance across all sentence lengths.
This lower performance is especially pronounced for sentences exceeding 250 characters, underscoring the limitations of using the \segmenters without capping.

Accordingly, we prepare the \lcm training  data with \sat Capped. 
We discuss in \Cref{app:data:technic} technical and engineering challenges faced when handling large amounts of \sonar embeddings.

\subsection{\LCM variants}
\label{sec:archi:lcm}
The design of the \lcm is driven by the need to conditionally generate a continuous sentence embedding. 
This obviously contrasts with how current \llms work, \ie estimating a probability distribution over a vocabulary of discrete tokens.
A straightforward way of solving the task is to train a transformer model to generate an embedding with the objective of minimizing the MSE loss (see \Cref{sec:arch:base}). 
However, a given context may have many plausible, yet semantically different, continuations.
The model should thus be able to learn a conditional probability distribution over the continuous embedding of the next sentence.

There is a large body of work in computer vision aiming to learn such conditional probability distributions over continuous data \citep{dhariwal2021diffusion,rombach_image_diffusion}. 
Models like Dall-E 3 \citep{dalle3} or Imagen Video \citep{ho2022imagenvideo} use a diffusion process to generate an image or video from a text prompt. 
Many different real images may satisfy the same input prompt, hence the model has to learn a probability distribution over continuous pixel data.
This motivates the exploration of diffusion models for sentence embedding generation.
Two variants are presented in \Cref{sec:arch:interleaved,sec:arch:twotower}.
Another prevalent take on continuous data generation consists of quantizing said data to ultimately model with discrete units; we explore \lcm modeling with quantization in \Cref{sec:arch:quantlcm}.

\subsubsection{\mselcm}
\label{sec:arch:base}
\begin{figure}[!t]
    \centering
    \includegraphics[width=.8\linewidth]{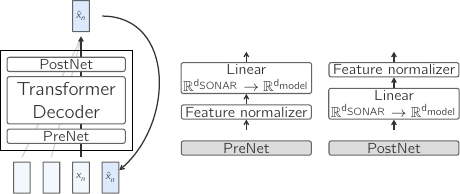}
    \caption{\textbf{The\mselcm.} Illustration of the \mselcm. At its core is a standard decoder-only Transformer surrounded with a $\prenet$ and a $\postnet$.}
    \label{fig:archi:baselcm}
\end{figure}

Our baseline architecture for next-concept prediction is a standard decoder-only Transformer that transduces a sequence of preceding concepts (read sentence embeddings) into a sequence of future ones. 
As illustrated in \Cref{fig:archi:baselcm}, the \mselcm is equipped with a ``$\postnet$'' and a ``$\prenet$''. The $\prenet$ normalizes the input \sonar embeddings and maps them to the model's hidden dimension $\modeldim$.
\begin{align}
\prenet(\rvx) &= \normalize(\rvx) \rmW_\text{pre}^t + \rvb_\text{pre}, \\
\postnet(\rvx) &= \denormalize\left(\rvx \rmW_\text{post}^t + \rvb_\text{post}\right), \\
\end{align}
where $\rmW_\text{post}\in\mathbb R^{\sonardim \times \modeldim}$, 
$\rvb_\text{post} \in \mathbb R^{\sonardim}$,
$\rmW_\text{pre}\in\mathbb R^{\modeldim \times \sonardim}$ and
$\rvb_\text{pre} \in \mathbb R^{\modeldim}$.

In order to learn the maps ``$\normalize$'' and its inverse ``$\denormalize$'' we fit a robust scaler to a set of randomly sampled \sonar vectors from different corpora and domains of text data. This scaler removes the median statistics and scales the data according to the interquartile range (IQR).
\begin{align}
\normalize(\rvx) = \frac{\rvx - \rvmu}{\rvsigma}, \quad
\denormalize(\rvx) = \rvmu + \rvsigma \rvx.
\label{eq:sonar:normalizer}
\end{align}

The \mselcm is trained on the semi-supervised task of next concept prediction, that is, the model predicts the next concept $\hat\rvx_n$ and its parameters $\rvtheta$ are optimized to regress the ground truth next concept ($\rvx_n$).
\begin{align}
  \hat\rvx_n =  f(\rvx_{<n}; \rvtheta),\quad \mse(\hat\rvx_n, \rvx_n) = \|\hat \rvx_n - \rvx_n\|^2.
\end{align}
Given a data distribution $\rq$ of documents (sequences of concepts), the training loss is evaluated as:
\begin{align}
\mathcal L_\mselcm(\rvtheta) = \mathbb E_{\rvx\sim q}\Big[\sum_{n=1}^{|\rvx|} \mse\left(f(\rvx_{<n}; \rvtheta), \rvx_n\right)\Big].
\end{align}

In order to enable the generation of variable length documents at inference time, we suffix training documents with the sentence ``End of text.''. Similar to any sentence in the document, this special suffix will be encoded with \sonar. This means that $\rvx_{|\rvx|} = \overrightarrow{\text{eot}} \defeq \encode(\text{"End of text."})$. During inference, we implement two main early stopping mechanisms: the first one measures the similarity of the generated embedding $\hat\rvx_n$ to $\overrightarrow{\text{eot}}$ and stops if the cosine similarity exceeds a threshold $s_\text{eot}$. The second mechanism compares the newly generated embedding $\hat\rvx_n$ to the previous generation $\hat\rvx_{n-1}$ and stops if their cosine similarity is higher than a threshold $s_\text{prev}$. We set both $s_\text{eot}$ and $s_\text{prev}$ to 0.9.

\subsubsection{Diffusion-based \lcm}
\label{sec:arch:diffsion:intro}

Diffusion-based \lcms are generative latent variable models that learn a model distribution $\ptheta$ approximating a data distribution $\rq$.
Similar to the \mselcm, we model the diffusion \lcms as auto-regressive models that generate concepts in a document, one at a time.
The model distribution is thus expressed at each position $n$ of the sequence as
$\ptheta(\diffx{}{n} | \diffx{}{<n})$ \ie the generation of the next concept is conditioned on the preceding context. 

In what follows we use a superscript for the denoising/diffusion step ($t\in[0, 1]$) and a subscript ($n$) for indexing the sequence of concepts. 
We simplify for a given $n$ the conditional model distribution 
$\ptheta(\diffx{0}{n} | \diffx{0}{<n})$
as 
$\ptheta(\diffx{0}{})$,
and the conditional data distribution 
$\rq(\diffx{0}{n}|\diffx{0}{<n})$ 
as 
$\rq(\diffx{0}{})$.

Diffusion models involve two processes: a \emph{forward} noising process and a \emph{reverse} denoising process~\citep{DDPM,DDIM}:
\paragraph{Forward process and noise schedule}
The forward process is a Gaussian diffusion process characterized by the marginal distribution $\rq(\diffx{t}{}|\diffx{0}{})$, given for every timestep $t \in [0,1]$ as:
\begin{align}
    \rq(\diffx{t}{}|\diffx{0}{}) \defeq \mathcal N(\alpha_t\diffx{0}{}, \sigma_t^2\rmI).
\end{align}
With the reparameterization trick, we can sample from this marginal distribution via:
\begin{align}
    \diffx{t}{} = \alpha_t\diffx{0}{} + \sigma_t\rvepsilon\quad\text{where } \rvepsilon\sim\mathcal N(\rvzero, \rmI)
    \label{eq:noising:reparam}
\end{align}
We use a variance-preserving forward process~\citep{karras2022elucidating} for which we have:
\begin{align}
    \alpha_t^2 = \sigmoid(\lambda_t), \quad
    \quad \sigma_t^2 = \sigmoid(-\lambda_t) = 1 - \sigmoid(\lambda_t), \quad
    \quad \lambda_t = \log\left({\alpha_t^2}/{\sigma_t^2}\right),
\end{align}
where $\lambda_t$ is the log signal-to-noise ratio (log-SNR) for timestep $t$.

The noise schedule is a strictly monotonically decreasing function $f_\lambda$ that maps from the timestep $t\in[0,1]$ to a log-SNR level: $\lambda_t = f_\lambda(t)$. 

It is common in previous work to also define the noise schedule based on a discrete variance schedule $(\beta_0, \ldots, \beta_T)$.
This stems from the formulation of the forward process as a discrete-time Markov chain that gradually adds Gaussian noise to the data according to said variance schedule:
\begin{align}
    \rq(\diffx{1\ldots \rT}{} | \diffx{0}{}) \defeq \prod_{t=1}^\rT \rq(\diffx{t}{} | \diffx{t-1}{}),\quad 
    \rq(\diffx{t}{} | \diffx{t-1}{}) \defeq \mathcal N(\diffx{t}{};\sqrt{1-\beta_t}\diffx{t-1}{},\beta_t\rmI),
    \label{eqn:schedule:discrete-distrib}
\end{align}
where to simplify the notation, $\diffx{t}{}$ is short for $\diffx{t/T}{}$ as the timesteps are now discretized.

From the variance schedule $(\beta_t)_t$, the noise schedule can be expressed as:
\begin{align}
    \alpha_t^2 =\prod_{s=1}^t (1-\beta_s).
\end{align}

Following \citet{kingma2024understanding}, for any given noise schedule, we visualize the distribution over noise levels $p(\lambda) = -dt/d\lambda$ in order to characterize how much time we are spending at every noise level during training.

\begin{figure}[!t]
    \centering
    \includegraphics[width=.85\linewidth]{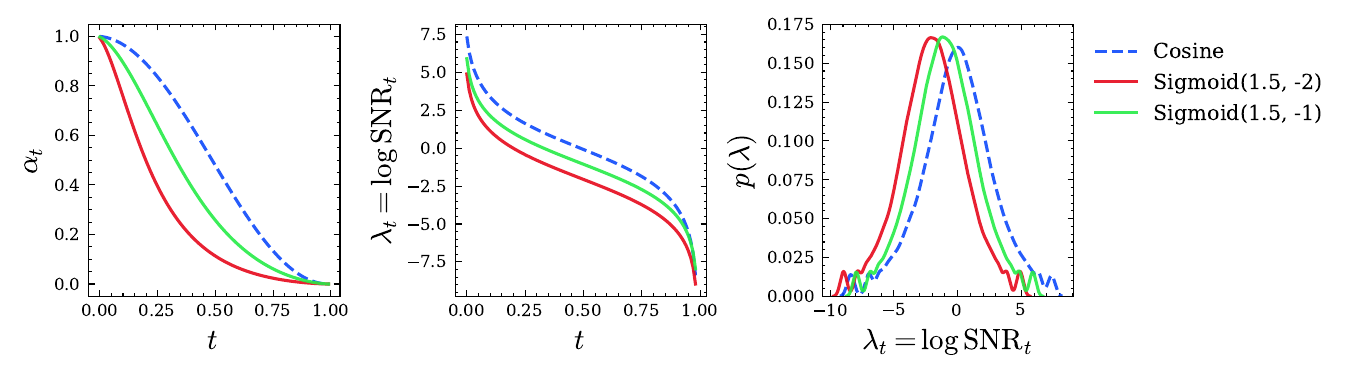}\\
    \includegraphics[width=.85\linewidth]{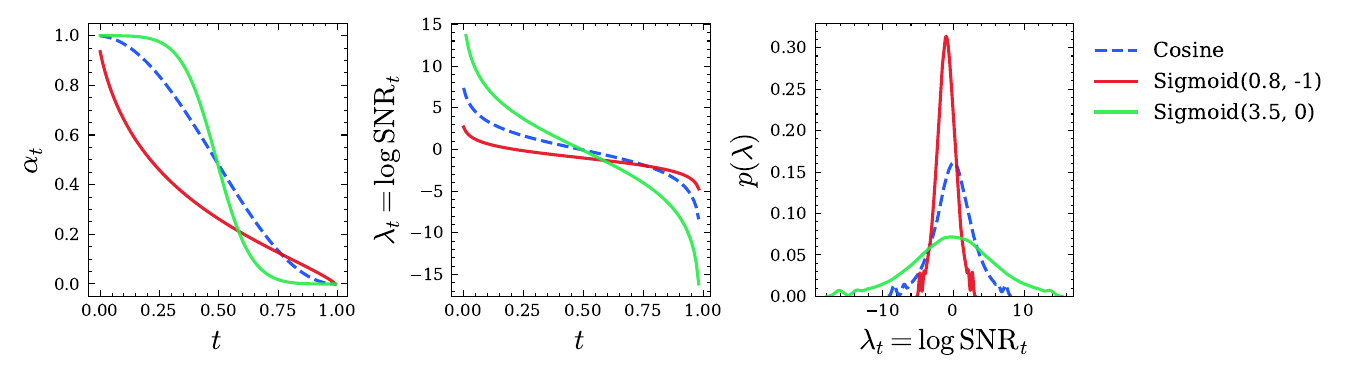}\\
    \includegraphics[width=.85\linewidth]{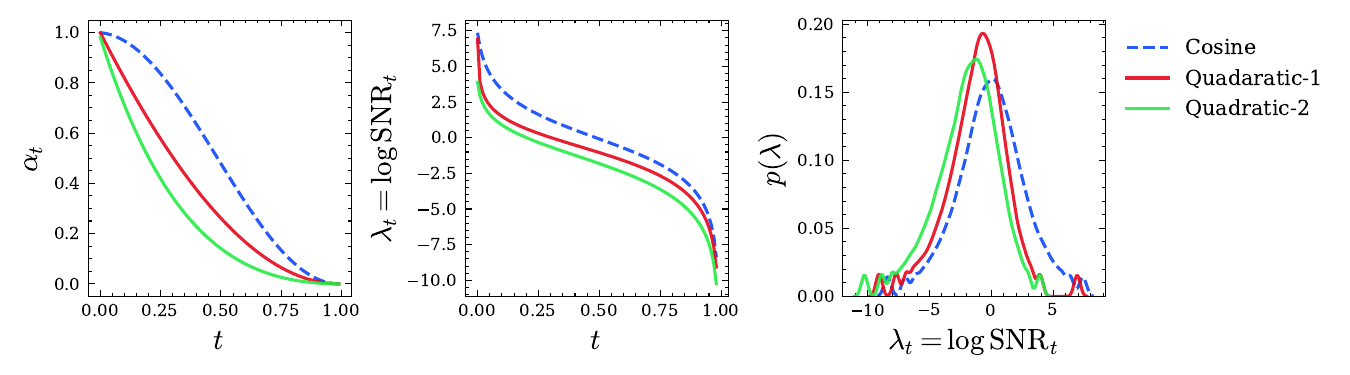}
    \caption{\textbf{Noise schedules.} Illustrations of the different noise schedules explored in this work. Our default schedule being cosine. Quadratic-1 is characterized by $(\beta_0=0.001, \beta_\rT=0.0012)$ and Quadratic-2 by $(\beta_0=0.02, \beta_\rT=0.022)$ For each schedule we visualize the curve of $(\alpha_t)_t$ (see \Cref{eq:noising:reparam}), the curve of the log-SNR and the associated distribution over noises levels $p(\lambda)$~\citep{kingma2024understanding}.}
    \label{fig:noise_schedules}
\end{figure}

In this work, we consider three types of noise schedules:
\begin{description}[style=unboxed,leftmargin=0cm]
\item[Cosine.] The schedule formulated in \citet{nichol2021improved} as:
\begin{align}
\alpha_t^2 = f(t)/f(0), \text{where } f(t) = \cos^2\left(\frac{t+s}{1+s}.\frac\pi2\right), \text{ where } s=0.008.
\end{align}
\item[Quadratic.] The schedule introduced in \citet{DDPM} where the variances $(\beta_t)_t$ are set to constants increasing quadratically from $\beta_0$ to $\beta_1$.
\begin{align}
\beta_{t/\rT} = \left(\sqrt{\beta_0} + \frac{t}{\rT}.\left(\sqrt{\beta_1} - \sqrt{\beta_0}\right)\right)^2.
\end{align}
\item[Sigmoid.] We introduce in this work, the \emph{sigmoid} schedule as a means to study the impact of the SNR distribution on the training of our models. The schedule is parametrized by two hyper-parameters ($\gamma, \delta)$ and is defined as:
\begin{align}
    \alpha_t^2 = f(t)/f(0),\text{ where } f(t) = \sigmoid\left(\delta - \gamma \logit(t)\right),
\end{align}
where ``$\sigmoid$'' is the sigmoid function $\sigmoid x\mapsto e^x / (e^x+1)$ 
and ``$\logit$'' its inverse function $\logit: x\mapsto \log(x/(1-x))$. The hyper-parameter $\gamma$ controls the scale of the log-SNR distribution $p(\lambda)$ and $\delta$ its center (see \Cref{fig:noise_schedules}).
\end{description}

In all our experiments, we follow \citet{lin2024common} and rescale the variance schedule $(\beta_1, \ldots \beta_T)$ to enforce zero terminal SNR \ie $\beta_\rT=1$.

\paragraph{Reverse process and objective function}
The joint distribution of the diffusion model $\ptheta(\diffx{0\ldots 1}{})$ is called the reverse process and is defined as a Markov chain with learned Gaussian transitions starting at $\rp(\diffx{1}{})=\mathcal N(\rvzero, \rmI)$. In its discretized form:
\begin{align}
    \ptheta(\diffx{0:\rT}{}) &\defeq \rp(\diffx{\rT}{})\prod_{t=1}^\rT \ptheta(\diffx{t-1}{}|\diffx{t}{}), \qquad
    \ptheta(\diffx{t-1}{}|\diffx{t}{}) \defeq \mathcal N(\diffx{t-1}{}; \rvmu_\rvtheta(\diffx{t}{}, t), \rmSigma_\rvtheta(\diffx{t}{}, t)),
\end{align}
where $\rvmu_\rvtheta$ and $\rmSigma$ are predicted statistics. $\rmSigma$ is set to to the constant $\sigma_t^2\rmI$ (matching the transitions of the forward process).
$\rvmu_\rvtheta$ can be decomposed into a linear combination of $\diffx{t-1}{}$ and a noise approximation model $\rvepsilon_\rvtheta$. This prediction method is dubbed $\rvepsilon$-prediction~\citep{DDPM,nichol2021improved,glide}. In this work we adopt $\diffx{0}{}$-prediction \ie we predict the noiseless state and optimize the simple reconstruction loss:
\begin{align}
    \loss(\rvtheta) \defeq \mathbb E_{t\sim\mathcal U(0,1)} \big[\omega(t) \loss(t, \rvtheta)\big],\quad
    \loss(t, \rvtheta) \defeq \mathbb E_{\diffx{0}{}, \rvepsilon} \Bigl[\big\| \diffx{0}{} - \rvmu_\rvtheta(\alpha_t\diffx{0}{} + \sigma_t\rvepsilon, t) \big\|_2^2\Bigr].
\end{align}
Different weighting strategies for the reconstruction loss were proposed in the literature~\citep{DDPM,maxsnr,minsnr}. In this work, we default to the simple reconstruction loss ($\omega(t)=1,\;\forall t)$ and we experiment with a clamped-SNR weighting strategy:
\begin{align}
    \omega(t) = \max(\min(\exp(\lambda_t), \lambda_{\max}), \lambda_{\min}), \, \lambda_t = \log(\alpha_t^2/\sigma_t^2),
\end{align}
which is a generalization of \citet{maxsnr}'s truncated-SNR weighting and \citet{minsnr}'s min-SNR strategy where the SNR is clamped between a min- and max-value $\lambda_{\min}$ and $\lambda_{\max}$.

Additionally, we consider a weighting strategy that factors in the quality of the sample $\diffx{0}{}$. We use as sample weight a scalar $\omega(\diffx{0}{})\in [0,1]$ correlated with the sample's fragility score \ie how easy it is to reconstruct a noised sample (see \Cref{sec:analysis:fragility}). Fragile samples will be assigned a smaller weight and thus contribute less to the objective function.
\begin{align}
\loss_{\fragility}(\rvtheta) &\defeq \mathbb E_{t\sim\mathcal U(0,1), \diffx{0}{}, \rvepsilon} 
\Bigl[\omega(\diffx{0}{})~\big\| \diffx{0}{} - \rvmu_\rvtheta(\alpha_t\diffx{0}{} + \sigma_t\rvepsilon, t) \big\|_2^2\Bigr],
\\
\omega(\diffx{0}{}) &= \sigmoid(a ~\fragility(\diffx{0}{}) + b),\label{eq:arch:loss:fragility}
\end{align}
where $a < 0$ and $b$ are hyper-parameters to tune.

\paragraph{Classifier-free diffusion guidance for the \lcm}
\label{sec:cfg}
Classifier-free diffusion guidance~\citep{ho2022classifier} consists of jointly training a conditional and an unconditional diffusion model. The resulting conditional and unconditional score
estimates are combined at inference time to achieve a trade-off between sample quality and diversity. This combined score is defined as follows:
\begin{align}
    \nabla_x \log_{\gamma} p(x|y) = (1-\gamma)\nabla_x \log p(x) + \gamma \nabla_x \log p(x|y),
\end{align}
where $\rvy$ is the conditioning variable, in our case the sequence of preceding embeddings $(\diffx{}{1}, \ldots \diffx{}{n-1})$ when denoising $\diffx{}{n}$.

The hyper-parameter $\gamma$ controls the contribution of the conditional score; For $\gamma= 0$, this is equivalent to an unconditional model, and for $\gamma = 1$, it is a fully conditional model. In practice for vision models, $\gamma$ is set to a value greater than 1, thus amplifying the signal from the conditioning model.

\paragraph{Inference}

At inference time, the reverse process is applied.
$\diffx{\rT}{}$ is obtained by sampling a random noise from $\rp(\diffx{\rT}{})=\gaussian(\rvzero, \rmI)$, and is then iteratively denoised by taking steps in the direction of the score function (\ie the direction in which the log-likelihood increases fastest).
Additional noise is added during the process in order to avoid falling down into modes of the distribution. 

Practically, we start from $\diffx{\rT}{} \sim\gaussian(\rvzero, \sigmainit^2\rmI)$ and find that the quality of the sampled output is sensitive to the initial noise scale $\sigmainit$. 

Although we train the model on a large number of discretized timesteps, \eg $\rT{=}100$, we only generate with a smaller number of steps, \eg $\rS{=}40$, at inference via accelerated generation processes~\citep{DDIM}. We select the sample steps following the trailing method of ~\citet{lu2022dpm} as it is found to be more efficient for smaller steps $\rS$~\citep{lin2024common}. That is we generate along the sampled steps
$(\tau_1, \ldots \tau_\rS)= \textrm{round}(\textrm{flip}(\textrm{arange}(\rT, 0, -\rT/\rS)))$.
During inference, we employ the classifier-free guidance rescaling technique of~\citet{lin2024common} proven to alleviate the image over-exposure problem encountered in image synthesis diffusion models as the terminal SNR approaches zero. We denote with $\guidance$ and $\rsguidance$ the guidance scale and guidance rescale factors used at inference.

Following~\citet{ning2023elucidating}, we perform Epsilon-scaling at inference time as it is shown to alleviate the exposure bias problem in diffusion models. In its simplified version, this is a training-free method that consists of scaling down the over-predicted magnitude of error by a scalar $\epscaling$.

We describe in \Cref{sec:arch:interleaved} and \Cref{sec:arch:twotower} two variants of diffusion \lcm: \interleaved and \twotower.

\begin{figure}[!htb]
    \centering
    \includegraphics[width=.9\linewidth]{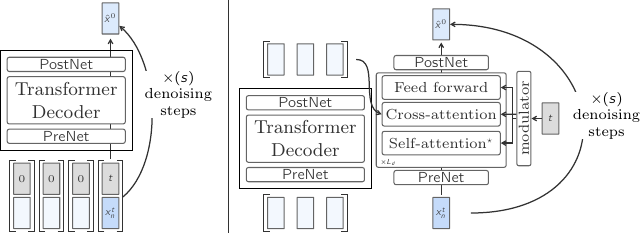}
    \caption{\textbf{Inference with diffusion-based \lcms.} In the left-hand side, an illustration of the \interleaved \lcm and on the right-hand side an illustration of the \twotower \lcm.
    }
    \label{fig:archi:difflcm}
\end{figure}

\begin{figure}[!htbp]
    \centering
    \includegraphics[width=.25\linewidth]{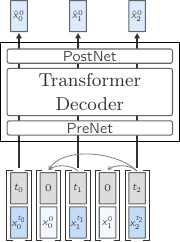}
    \caption{\textbf{Training of \interleaved diffusion \lcm.} Interleaving the clean and noisy embeddings and sampling different diffusion timesteps allows for efficient training.
    }
    \label{fig:archi:interleaved_training}
\end{figure}

\subsubsection{\interleaved Diffusion \lcm}
\label{sec:arch:interleaved}

This model, depicted in the left panel of \Cref{fig:archi:difflcm}, consists of a single transformer backbone whose task is to predict the clean next sentence embedding $\diffx{0}{n}$ given a noisy input $\diffx{t}{n}$, conditioned on previous clean sentence embeddings $\diffx{0}{<n}$.
During training, self-attention can be dropped with a certain probability
for unconditional training. 
This enables classifier-free guidance at inference time (see \Cref{sec:cfg} for details).

Each input embedding is concatenated with the corresponding diffusion timestep embedding. 
The learned position embeddings are added to the input vectors prior to being fed to \lcm.
The backbone utilizes a causal multi-head self-attention.

For efficient training, the model is trained to predict each and every sentence in a document at once.
As depicted in \Cref{fig:archi:interleaved_training}, during the diffusion process, the model attends to the clean sentences in the context using causal multi-head attention layers.
The input is specially prepared by interleaving the noisy (blue) and clean (light blue) sentence embeddings, and the attention mask is prepared accordingly to only attend to the clean sentence embeddings (gray arrows).

\subsubsection{\twotower Diffusion \lcm}
\label{sec:arch:twotower}
This model, depicted in the right panel of \Cref{fig:archi:difflcm}, separates the encoding of the preceding context from the diffusion of the next embedding.
A first model, labeled \emph{\ctxenc}, takes as input the context vectors $x_{<n}$ and encodes them causally \ie we apply a decoder-only Transformer with causal self-attention.
The outputs of the \ctxenc are then fed to a second model dubbed \emph{\denoiser}, which predicts the clean next sentence embedding $\diffx{0}{n}$ by iteratively denoising the latent $\diffx{1}{n} \sim \mathcal N(\rvzero, \rmI)$. 
The \denoiser consists of a stack of Transformer blocks with cross-attention block to attend over the encoded context.
Both the \denoiser and the \ctxenc share the same Transformer hidden dimension $\modeldim$.
Each block of each Transformer layer in the \denoiser (including the cross-attention layer) is modulated with adaptive layer norm ($\adaln$, ~\citet{perez2018film, peebles2023scalable}). The $\adaln$ modulator of \twotower regresses channel-wise scale ($\rvgamma$), shift ($\rvbeta$) and residual gates ($\rvalpha$) from the embedding of the current diffusion timestep $t$.

\begin{align}
    [\rvbeta, \rvgamma, \rvalpha] & = \silu(\tembed(t)) \rmW^t + \rvb, 
    \label{eq:twotower:adaln:regress}\\
    \rvy & = \rvx + \rvalpha \,\tfblock((1+\rvgamma)\,\rvx + \rvbeta),\label{eq:twotower:adalan:modulate}
\end{align}

Following \citet{peebles2023scalable} and \citet{goyal2017accurate} we initialize each residual block in a Transformer layer (``$\tfblock$'') with the identity function via initializing $\rmW$ and $b$ in \Cref{eq:twotower:adaln:regress} to zero.
The diffusion timestep $t$ is embedded using a 256-dimensional frequency embedding~\citep{dhariwal2021diffusion,peebles2023scalable} followed by a two-layer MLP with $\silu$ as activation function. ``$\tembed$'' maps to the \denoiser's hidden dimension $\modeldim$. The self-attention layers in the \denoiser do only attend to the current position \ie we do not attend to the preceding noised context. The self-attention layers were kept for consistency with a standard Transformer block and for the possible extension of denoising multiple vectors at once.

\begin{figure}[!t]
    \centering
    \includegraphics[width=.8\linewidth]{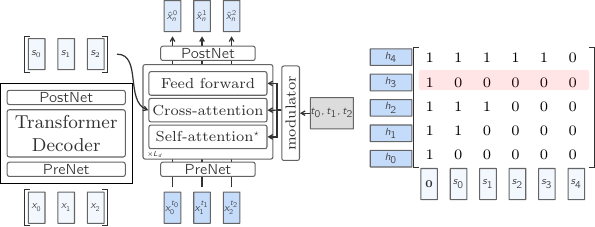}
    \caption{\textbf{Training \twotower diffusion \lcm.} On the left panel, a \twotower forward pass in training time in order to denoise multiple embeddings in parallel. On the right side panel a visualization of the \denoiser's cross-attention masks with the red highlighted row signaling a sample dropped to train the \denoiser unconditionally. $(h_1, \ldots, h_4)$ denotes the sequence of intermediate representations in the \denoiser right before the cross-attention layer.}
    \label{fig:archi:twotower:attn}
\end{figure}

\vspace{-2mm} %
\subparagraph{\twotower training.}
At training time, \twotower's parameters are optimized for the next-sentence prediction task on unsupervised sequences of embeddings. The causal embeddings from the \ctxenc are shifted by one position in the \denoiser and a causal mask is used in its cross-attention layers. A zero vector is prepended to the context vectors to enable the prediction of the first position in the sequence (see \Cref{fig:archi:twotower:attn}).
To train the model both conditionally and unconditionally in preparation for inference with classifier-free guidance scaling, we drop random rows from the cross-attention mask with a rate of $\rp_\text{cfg}$ and denoise the corresponding positions with only the zero vector as context.

\subsubsection{Quantized \lcm}
\label{sec:arch:quantlcm}
Two major approaches currently stand to deal with continuous data generation in the image or speech generation fields: one is diffusion modeling, the other is learning quantization of the data before modeling on top of these discrete units.

In addition, the text modality remains discrete, and despite dealing with continuous representations in the SONAR space, all possible text sentences (of less than a given number of characters) are a cloud of points rather than a real continuous distribution in the SONAR space. 
These considerations motivate the exploration of quantization of SONAR representations and then modeling on these discrete units to address the next sentence prediction task. Finally, following such an approach enables the natural use of temperature, top-p or top-k sampling, to control the level of randomness and diversity in the sampling of the next sentence representation.

In this section, we learn residual quantizers for the SONAR space, and then build a Quantized \LCM based on these discrete units. We tried to come up with an architecture as close as the diffusion \lcm models, to be able to compare approaches.

\paragraph{Quantization of SONAR space.} We use Residual Vector Quantization (RVQ; \citet{zeghidour2021soundstream}) as a coarse-to-fine quantization technique to discretize SONAR representations. Vector quantization maps continuous input embeddings to the nearest entry in a learnt codebook. RVQ iteratively quantize residual errors from previous quantizations using additional codebook for each iteration. We use FAISS implementation~\citep{douze2024faiss}
which performs iterative k-means clustering of residuals. We use the Improved Residual Vector Quantization (IRVQ) method from \citet{liu2015improved}, with a beam size of 1 for memory efficiency. We trained the RVQ codebooks on 15 million English sentences extracted from Common Crawl using $n_\text{codebooks} = 64$ number of quantizers with $n_\text{units-per-codebook}=8192$ units per codebook. 

\begin{figure}[!ht]
    \centering
    \includegraphics[width=0.4\linewidth]{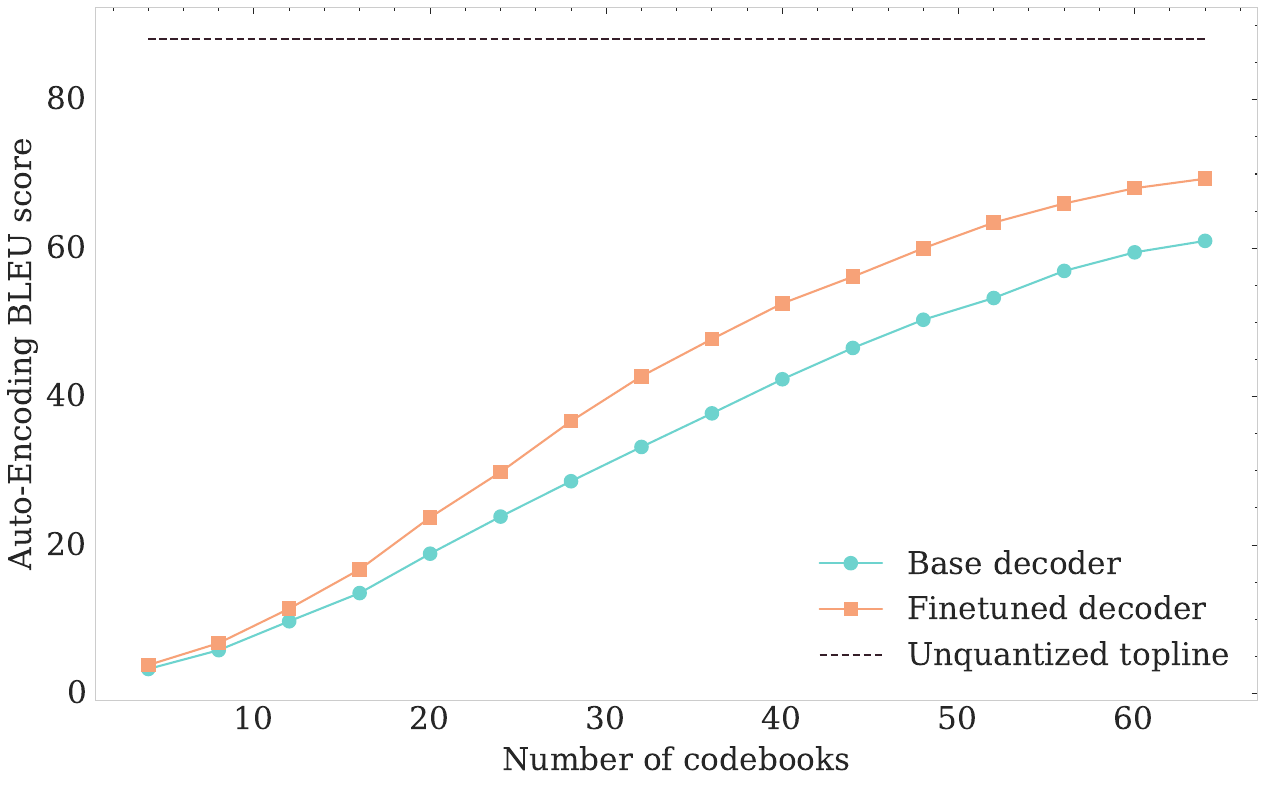}
    \caption{Auto-encoding BLEU scores on FLORES devtest set, encoding sentences with SONAR encoder, quantizing with a varying number of codebooks, dequantizing and decoding with SONAR decoder.}
    \label{fig:auto_bleu_quant}
\end{figure}

One property of RVQ is that the cumulative sum of centroid embeddings of the first codebooks are an intermediate coarse approximation of input SONAR vectors. In that way, we can report the evolution of auto-encoding BLEU scores with the increasing number of codebooks used to quantize SONAR embeddings, before using the SONAR text decoder to decode quantized embeddings. We notice in \autoref{fig:auto_bleu_quant} that auto-encoding BLEU consistently improves as the number of codebooks increases , reaching around 70\% of the auto-encoding BLEU score achieved with continuous SONAR embeddings, when using all 64 codebooks. %

\

\paragraph{Finetuning the SONAR decoder on quantized representations.}
We fine-tuned SONAR decoder on quantized representations to adjust it for the space created by the quantizers on 1.2M English sentences.
To make the decoder more robust against residual representations from intermediate codebooks, we randomly select a codebook number $k \in \bigl[\frac{2}{3} \cdot n_\text{codebooks}, n_\text{codebooks}\bigl]$ during fine-tuning, with probability $p = 0.3$, and use the quantized representation with codebooks up to $k$.
\Cref{fig:auto_bleu_quant} shows the improvement in auto-encoding performance when the decoder is adapted to quantized representations.

\paragraph{\qlcm architecture.}
In the same spirit of diffusion \lcm, we aim at coarse-to-fine generation of \sonar embeddings conditioned on left-context sentences. However, we do not follow a denoising task as in diffusion modeling, but an iterative generation of \sonar embeddings based on \textit{intermediate quantized representations} instead. In order to generate a \sonar embedding conditioned on left-context sentences, the \qlcm model starts with the \textit{intermediate representation} as a  vector filled with zeros. We iteratively add to this \textit{intermediate representation} the predicted residual centroid embeddings. In that way, the predicted \sonar embeddings are iteratively refined based on the growing cumulative sum of centroid embeddings of first codebooks, until all codebooks have been seen. We used the \interleaved architecture for \qlcm experiments even though it could be trained with \twotower architecture too. Compared to the diffusion \lcm, noisy input representations are replaced with \textit{intermediate quantized representations} and diffusion timestep embeddings as input are replaced by codebook index embeddings. 

\paragraph{Discrete targets.}
Following previous work on modeling discrete units from residual quantizers \citep{valle,audiopalm,lee2022autoregressive}, a Quant-LCM can be trained to predict the unit from the next codebook, parameterized with a softmax output layer. For parameter efficiency, we do not use $n_\text{codebooks} \cdot n_\text{units-per-codebook}$  unique indices as discrete targets which would imply $n_\text{codebooks} \cdot n_\text{units-per-codebook}$ output dimensions, but only $n_\text{units-per-codebook}$ output dimensions while inputting the information of the codebook index to the model. At training time, similarly to diffusion \lcm training, we randomly sample codebook index $k$ between 1 and  $n_\text{codebooks}$, and compute the cumulative sum of centroid embeddings of the first $k {-}1$ codebooks as input. We use the unit from codebook $k$ of the target embedding as target index for cross entropy loss computation. At inference time, we iteratively predict the unit from the next codebook, get the corresponding centroid embedding and add it to the current \textit{intermediate representation} as additional predicted residual embedding. Finally, we also enable classifier-free guidance on logits at inference time~\citep{gafni2022make}
by randomly dropping left-context conditioning during training as previously described in \Cref{sec:arch:interleaved}. This modeling approach with discrete targets is dubbed \qlcmd in the following sections. The improved SONAR decoder for quantized representations is used to bridge the compression gap coming from SONAR quantization in following ablation studies when using \qlcmd.

\paragraph{Continuous targets.} 
We also explored a modeling approach that predicts continuous target SONAR vectors based on left-context sentences and intermediate quantized representation of the target vector, minimizing the Mean Squared Error between prediction and target embeddings. At inference time, we can either iteratively add the closest centroid embedding based on the predicted residual $\hat{\rvr}$ or sample a centroid $\rvc_i$ from the following distribution: 
\begin{align}
\rp(\rvc_i| \hat\rvr) = \dfrac{
e^{-\beta \cdot \| \rvc_i - \hat{\rvr} \|_2}
}{
\sum_{k}{e^{-\beta \cdot \| \rvc_k - \hat{\rvr} \|_2}}},
\end{align}
where $\beta$ is a temperature hyper-parameter. This modeling approach with continuous targets is denoted with \qlcmc in the following sections.

\subsection{Ablations}
\label{sec:archi:lcm:ablations}
In this section, we delineate the ablations experiments conducted to evaluate the aforementioned \lcm designs.
We compare all the variants of \lcms introduced above, namely, \mselcm, \interleaved, \twotower and \qlcm.

\subsubsection{Experimental setup}\label{sec:archi:lcm:ablation:setup}

For our ablation study and for the sake of reproducibility, we pre-train our models on the \fineweb dataset~\citep{fineweb-edu}.
All models are configured to have approximately 1.6B trainable parameters and are pre-trained on Meta’s Research Super Cluster (RSC,~\citet{rsc}) for 250k optimization steps spanning 32 A100 GPUs with a total batch size of 229k concepts.

\paragraph{Models architectures.}
The \mselcm has 32 layers and a model dimension $\modeldim=2048$ with 16 attention heads. It uses rotary position embeddings~(RoPE, \citet{su2024roformer}), applies pre-normalization using RMSNorm~\citep{zhang2019root}, uses the SwiGLU activation function~\citep{shazeer2020glu} and is trained with a dropout rate of $\rp{=}0.1$.

The \interleaved diffusion \lcm is made of 32 transformer blocks, 
each made of a self-attention layer with 32 attention heads and followed by a feed-forward neural network with inner size 8192. 
It has a dimension $\modeldim$ of 2048 and uses learned position embeddings.
The noise scheduler is set with $\rT{=}100$ diffusion timesteps.
During training, self-attention is dropped with a probability of 0.15 for unconditional training, enabling classifier-free guidance at inference time.

The \twotower diffusion \lcm has 5 layers in its \ctxenc and 13 layers in its \denoiser.
Similar to the \mselcm, it has 16 attention heads, a model dimension $\modeldim=2048$, and uses SwiGLU activations and RMSNorm in both \ctxenc and \denoiser. The \ctxenc uses RoPE for embedding positions whereas the \denoiser is without positional embeddings. We use by default the cosine noise schedule with $\rT{=}100$ and train with a dropout rate of $\rp{=}0.1$. For training the model unconditionally we use a cross-attention mask dropout of rate 0.15 (see \Cref{sec:arch:twotower}). The pre-training documents are wrapped at 128 sentences.
Unless otherwise mentioned we decode with $\rS{=}40$ sample steps with a guidance scale $\guidance=3$, a guidance rescaling factor of $\rsguidance=0.7$, an initial noise scale $\sigmainit=0.6$ and epsilon-scaling with $\epscaling=1.00045$.

The \qlcm follows exactly the same architecture as the \interleaved diffusion \lcm, except for \qlcmd which differs only by its output dimension which is set to $n_\text{units-per-codebook}=8192$ for softmax computation. For single sentence prediction tasks, we use $\text{top}_k=1$ and $\guidance=2$  for \qlcmd and $\text{top}_k=1$ with $\guidance=3$ for \qlcmc, while for multi-sentence generation tasks we used temperature of 1,  $\text{top}_k=3$, $\guidance=1$, for \qlcmd and temperature of 0.005, $\text{top}_k=5$, $\guidance=1.5$ for \qlcmc, as higher guidance or lower temperature setups led to repeated generated sentences.

\begin{table}[!htb]
\centering
\begin{tabular}{@{}lrrrrrrrr@{}}
\toprule
 & & \multicolumn{3}{c}{\#\llamatwo Tokens} & \multicolumn{3}{c}{\#Sentences} & \multirow{2}*{\makecell[c]{Total\\sentences}} \\
\cmidrule(lr){3-5}  \cmidrule(lr){6-8} 
Dataset & \#Docs & Q1 & Q2 & Q3 & Q1 & Q2 & Q3 &  \\
\midrule
\rocstories (dev) & 2000 & 48 &	57 & 64 & 5 & 5	& 5 & 10K\\
\rocstories (test) & 1871 & 50 & 56	& 62 & 5 & 5	& 5 & 9.4K \\

\cfour (dev) & 1000 & 136 & 288 & 577 & 6 & 12 & 24 & 20.6K \\
\cfour (test) & 1000 & 133 & 282 & 599 & 6 & 11 & 25 & 21.9K \\

\wikipedia (dev) & 1000 & 146 & 332 & 736 & 5 & 10 & 23 & 21.1K \\
\wikipedia (test) & 1000 & 147 & 312 & 673 & 5 & 9 & 21 & 19.1K \\

\gutenberg (dev) & 55 & 10297 &	15934 & 22259 & 328 & 530	& 687 & 216.2K \\
\gutenberg (test) & 61 & 10752 & 15204 & 23414 & 325 & 457 & 735 & 562.2K \\
\bottomrule
\end{tabular}
\caption{\textbf{Statistics of the pre-training evaluation datasets.} For each subset we report the number of documents, the total number of sentences and document lengths quartiles in sentences and in \llamatwo tokens for reference.}
\label{tab:ablation:eval_data_statistics}
\end{table}

\paragraph{Pre-training evaluation.} 
Pre-trained token-level language models are typically evaluated with perplexity: a measure of how well each next token is predicted given a teacher-forced (\ie ground truth) prefix of the document. In a similar spirit, we evaluate pre-trained LCMs in a teacher-forced mode. But as they cannot produce the probability explicitly, we resort to a custom set of metrics of the quality of next sentence prediction.

Each pre-trained model is initially evaluated on the quality of its predicted next sentence $\hat\rvx_n$ given a ground truth context $\diffx{}{<n}$. Practically, for a given document $\rvx_{1:\rN}$, we run the LCM inference in teacher-forcing mode and evaluate the following metrics:
\begin{itemize}[style=unboxed,leftmargin=*]

\item \textbf{L2 distance} ($\ltwo$). Euclidean distance in the \sonar space between the predicted embedding $\hat \rvx_n$ and the ground truth continuation $\rvx_n$: $\ell_2 \defeq \|\hat \rvx_n - \rvx_n\|^2$.

\item \textbf{Round-trip L2 distance} ($\ltworound$). Euclidean distance in the \sonar space between the re-encoded sentence generated from the predicted embedding and the ground truth continuation $\rvx_n$, \[\ltworound \defeq \|\encode(\decode(\hat \rvx_n)) - \rvx_n\|^2.\]
Since an LCM can predict an embedding outside of the distribution of real embeddings (obtained by encoding natural sentences), the \sonar decoder might shift these embeddings to the nearest plausible embeddings subspace. The $\ltworound$ metric is introduced to capture the shift in embeddings after decoding them into text then re-embedding them again in the \sonar space. The more the generated embeddings are out-of-distribution, the higher the delta between $\ltworound$ and $\ltwo$ would be.

\item \textbf{Contrastive accuracy} ($\mseacc$). The ratio of embeddings in a batch that are further away (in terms of $\ltwo$) from the predicted embedding $\hat \rvx_n$ than the ground truth $\rvx_n$ (for each $n$, we exclude $\rvx_n$ and its two neighboring ground truth embeddings from the comparison). This metric naturally assigns higher penalty for large $\ltwo$ values in the regions with high density of sentence embeddings. 

\item \textbf{Paraphrasing} ($\paraphrasing$). The maximum cosine similarity ($\cossim$) between the generated embedding $\hat \rvx_n$ and the context embeddings $\rvx_{<n}$,
normalized by the score of the ground truth sentence. 
Thus, $\paraphrasing = \max\limits_{m<n} \cossim(\hat \rvx_n, \rvx_{m}) / \max\limits_{m<n} \cossim(\rvx_n, \rvx_{m})$.
The goal of this metric is to capture if the model is simply copying or paraphrasing a sentence from the context more ($>1$) or less ($<1$) than the reference sentence.

\item \textbf{Mutual information} ($\mutinfo$). This metric of text coherence evaluates the mutual information between the next predicted sentence $\hat \rs_n = \decode(\hat \rvx_n)$ and the previous $k=10$  ground truth sentences by computing the difference between the unconditional perplexity of $\hat s_n$ and its perplexity conditioned on the prompt:
\[ \mutinfo = \frac{1}{|\hat \rs_n|}\left( \log \rp_\text{LM}(\hat s_n) - \log \rp_\text{LM}(\hat \rs_n|\rs_{n-k:n-1}) \right).\]
We estimate the perplexity with a small language model, GPT-2 \citep{radford2019language}. We prepend a newline symbol to $\hat \rs_n$, so that a probability could be assigned to its first token, and we compute the average mutual information per-token by normalizing it with $|\hat \rs_n|$, the length of $\hat \rs_n$ in tokens. When averaging $\mutinfo$ over a dataset, $|\hat \rs_n|$ are used as weights.
\end{itemize}

\paragraph{Pre-training evaluation data.}
The pre-training evaluation is performed on sampled subsets from four corpora covering different domains: \rocstories~\citep{rocstories}, \cfour~\citep{c4}, \wikipedia (English Wikipedia dump) and \gutenberg.  
We sample two distinct subsets (dev and test) from each corpus, we use the dev split for tuning inference hyper-parameters and report the results on the test splits.
The statistics of the evaluation corpora are presented in \Cref{tab:ablation:eval_data_statistics}.

\begin{table}[hbtp!]
    \centering
    \begin{tabular}{llllll|lllll}
        \toprule
        \multirow{2}*{Model} 
        & \multicolumn{5}{c|}{\rocstories} 
        & \multicolumn{5}{c}{\cfour}  
        \\
        \cmidrule(lr){2-6} \cmidrule(lr){7-11}
         & $\ltwo$ & $\ltworound$ & $\paraphrasing$ & $\mseacc$ & $\mutinfo$ 
         & $\ltwo$ & $\ltworound$ & $\paraphrasing$ & $\mseacc$ & $\mutinfo$
         \\
        \midrule
        \mselcm & 0.177 & 0.237  & 1.847  & 72.4\% & 0.062 
        & 0.204 & 0.261 & 1.964  & 69.1\% & -0.105 \\
        
        \interleaved & 0.236 & 0.236 & 1.939 & 80.2\% & 0.977 
                     & 0.279 & 0.273 & 2.239 & 77.1\% & 1.110 \\
        
        \twotower & 0.233 & 0.231 & 2.088 & 80.6\% & 1.137
                  & 0.265 & 0.261 &  2.265 & 75.4\% & 1.134 \\
        
        \qlcmc & 0.236 & 0.237 & 1.683 & 76.0\% & 0.610 
         & 0.279 & 0.283 & 2.013 & 77.2\% & 0.715 \\
        
        \qlcmd  & 0.240 & 0.246 & 1.871 & 81.1\% & 0.682
        & 0.270 & 0.282 & 1.808 & 75.0\% & 0.359\\ 
        \bottomrule
    \end{tabular}
    \begin{tabular}{llllll|lllll}
        \toprule
        \multirow{2}*{Model} 
        & \multicolumn{5}{c|}{\wikipedia} 
        & \multicolumn{5}{c}{\gutenberg}    \\
         \cmidrule(lr){2-6} \cmidrule(lr){7-11}
         & $\ltwo$ & $\ltworound$ & $\paraphrasing$ & $\mseacc$ & $\mutinfo$ 
         & $\ltwo$ & $\ltworound$ & $\paraphrasing$ & $\mseacc$ & $\mutinfo$
         \\
        \midrule
        \mselcm & 0.229 & 0.283 & 1.770  & 69.6\% & 0.071 
        & 0.207 & 0.264 & 1.780 & 67.8\% & -0.184\\
        \interleaved & 0.324 & 0.311 & 2.087 & 80.9\% & 1.202 
                     & 0.284 & 0.281 & 2.051 & 75.1\% & 0.725 \\
       
        \twotower & 0.307 & 0.297 & 2.079 & 78.8\% & 1.307 &
        0.267 & 0.267 & 2.077 & 73.0\% & 0.684 \\
        \qlcmc  & 0.306 & 0.317 & 1.842 & 79.5\% & 0.744 & 
        0.269 & 0.281 & 1.774 & 72.1\% & 0.419 \\
        \qlcmd   & 0.295 & 0.311 & 1.592 & 76.0\% & 0.323 &
        0.276 & 0.290 & 1.599 & 72.0\% & 0.153\\
        \bottomrule
    \end{tabular}
    \caption{\textbf{Comparing architectures.} Pre-training evaluation results on the four select corpora. For each subset, we report $\ltwo$ (L2 distance in \sonar space), $\ltworound$ (round-trip L2 distance after decoding and re-encoding the generated embeddings), $\paraphrasing$ (similarity to preceding embeddings) and $\mseacc$ (contrastive accuracy)}
    \label{tab:ablation_results_nsp}
\end{table}

The results of the pre-training evaluation are presented in \Cref{tab:ablation_results_nsp}.

First, diffusion-based \lcm and \qlcm variants have similar $\ltwo$ and $\ltworound$ scores despite an important difference in their learning objectives. 
The only model that shows substantially lower $\ltwo$ score is the \mselcm. 
This is expected since \mselcm effectively optimizes $\ltwo$ score during training. 
Yet, $\ltworound$ score is not improved compared to other models.
This could be explained by the fact that when many plausible next sentence continuations are possible, \mselcm generates their average in SONAR space (instead of sampling one of plausible modes) which may not correspond to any relevant point in \sonar space.
This hypothesis is also highlighted by the poor \mselcm performance in term of $\mseacc$ and $\mutinfo$ scores.

We do not notice any significant difference in $\mseacc$ scores between diffusion \lcms and \qlcm variants. $\mutinfo$ scores, on the contrary, are consistently higher for diffusion-based models compared to \qlcm. At the same time, diffusion \lcms tend to paraphrase more the context in the generated embeddings, which also correlates with an increased $\mutinfo$ score.
Still, \qlcm variants significantly outperform \mselcm on $\mutinfo$ metric.
Now comparing the different variants, \qlcmc outperforms \qlcmd modeling variant: one hypothesis is that predicting codebook indices with cross-entropy loss is harder than $\mse$ objective where \qlcmc can more easily learn combination of left-context vectors for next sentence embedding.

For diffusion \lcms, we don't observe any consistent difference between \interleaved and \twotower when looking across all metrics and datasets. 
Note that overall, to tackle the next sentence prediction task in the SONAR space, diffusion-based methods give clearly better results compared to all other models.

\paragraph{Instruction-tuning evaluation.}
Subsequently, the pre-trained models are instruction-tuned on the stories subset of \cosmopedia~\citep{cosmopedia} and are evaluated on a held-out subset of \cosmopedia itself. We aim with this finetuning to evaluate the ability of the models to follow instructions and generate consistent stories. 

For the sake of comparison, we trained a small Llama~\citep{touvron2023llama} on the same training data (\fineweb) and finetuned it on \cosmopedia.
This model has 24 transformer layers, each with 16 attention heads and a model dimension of 2048 for a total of 1.4B parameters. This model will be referred to as \sllama.

We evaluate the following metrics:
\begin{itemize}[style=unboxed, leftmargin=*]
    \item \textbf{ROUGE-L} ($\rougel$). ROUGE-L (F-measure)~\citep{lin2004rouge} between the generated and reference stories.
    \item \textbf{Coherence} ($\coherence$). This reference-free metric is computed with a bidirectional transformer model fine-tuned by \citet{jwalapuram-etal-2022-rethinking} to assign higher scores to positive ``natural'' documents than to negative examples with permuted sentences. For reporting, we normalize it with a sigmoid (with a temperature 3.0, empirically set to make the scores of ``certainly incoherent'' documents close to 0 and those of ``certainly coherent'' documents close to 1). 
\end{itemize}

\begin{table}[!htb]
    \centering
    \begin{tabular}{lrr}
        \toprule
        Model & $\rougel\uparrow$ & $\coherence\uparrow$ \\
        \midrule
        \mselcm &  23.69 & 0.482 \\
        
        \interleaved & 33.40 & 0.968  \\
        
        \twotower &  33.64 & 0.938 \\

        \qlcmc & 30.87 & 0.847  \\
        \qlcmd & 28.01 & 0.704  \\
        \midrule
        \sllama & 34.88 & 0.984\\
        \bottomrule
    \end{tabular}
    \caption{\textbf{Comparing architectures.} Instruction-tuning evaluation results. For each model we score the generated stories on the held-out test prompts and report $\rougel$ (ROUGE-L) scores.}
    \label{tab:ablation_results_finetuning}
\end{table}

The scores in terms of $\rougel$ and $\coherence$ of the finetuning evaluation are presented in \Cref{tab:ablation_results_finetuning}.
Those quantitative results are in line with the pretraining evaluation ones. 
Both $\rougel$ and $\coherence$ scores correlate with the model ordering based from $\mutinfo$ scores, mainly \qlcm is outperformed by diffusion-based models, and both outperform \mselcm by a large margin. 

We also note that \sllama outperforms the \lcms on this downstream task on both metrics.
It is well known that \llms produce very fluent outputs, that explains the higher \rougellong score.
We also note that the \interleaved and \twotower produce coherent outputs, on par with the \sllama outputs.

\subsubsection{Importance of the diffusion inference hyper-parameters}
\begin{figure}
\centering
\includegraphics[width=\linewidth]{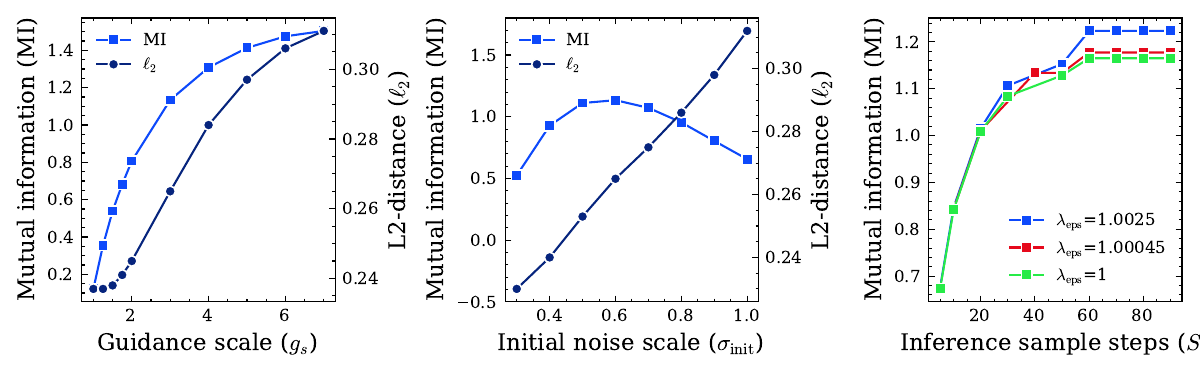}
\caption{\textbf{Importance of inference hyper-parameters.} The first panel shows the quality of the generated output measured with $\mutinfo$ and $\ltwo$ as we vary the guidance scale $\guidance$ with fixed $\sigmainit=0.6$ and $\rS=40$. The second panel varies the initial noise scale $\sigmainit$ with fixed guidance $\guidance=3$ and $\rS=40$. The third panel varies the inference steps $\rS$ while holding the guidance scale $\guidance=1.5$ and $\sigmainit=0.6$ fixed. We consider 3 values for $\epscaling$ to see the impact of epsilon-scaling in the regime of large inference steps.}
\label{fig:arch:ablation:inference}
\end{figure}

In this section we study the effect of different inference hyper-parameters on the quality of the generated text. To this end, we generate outputs for the \cfour test split with the \twotower \lcm model above, while varying the following hyper-parameters: the guidance scale $\guidance$, the initial noise scale $\sigmainit$, and the number of inference sample steps $\rS$. We score the generations following the same protocol above and report the results in \Cref{fig:arch:ablation:inference}. We note that as we increase the guidance scale, the mutual information between the prefix and the generated suffix increases, and so does paraphrasing as we are paying more attention to the conditioning context variables. In the opposite direction of the mutual information, the $\ltwo$ distance from the ground truth continuation increases as the model prefers to stay close to the prefix.   Regarding the initial noise scale, we observe that values between 0.5 and 0.7 achieve the best $\mutinfo$ score. In particular, the generated individual sentences are usually longer with a higher $\sigmainit$. The $\ltwo$ distance on the other hand does not reflect this trend. Lastly, we increase the number of inference steps and measure the mutual information ($\mutinfo$) of the generated texts. With more inference steps, we can improve the prefix-suffix mutual information, but there is diminishing returns to increasing the inference cost with little qualitative improvement.

\subsubsection{Studying the noise schedules}
In this section, we compare different \twotower diffusion \lcms trained with different noise schedules, namely:
\begin{description}[style=unboxed,leftmargin=0cm]
    \item[Cosine.] Our default cosine noise schedule.
    \item[Quadratic.] The first quadratic schedule (Quadratic-1) has $\beta_0=0.001$ and $\beta_T=0.0012$, whereas the second quadratic schedule (Quadratic-2) has $\beta_0=0.02$ and $\beta_T=0.022$.
    \item[Sigmoid.] Four different sigmoid schedules with with $(\alpha,\beta) \in \{(1.5, -1), (1.5, -2), (0.8, -1), (3.5, 0)\}$.
\end{description}
All schedules are set with the default $\rT{=}100$. For the exact description of each noise schedule refer to \Cref{sec:arch:diffsion:intro} and \Cref{fig:noise_schedules}.
We selected the quadratic schedule as a commonly used schedule for reference with two configurations, Quadratic-1 closer to the cosine schedule and Quadratic-2 with more weight given to lower log-SNR. The selected sigmoid schedules with $\delta=1.5$ are configured with $\gamma=-1$ and $\gamma=-2$, $\gamma=-2$ being slightly shifted to the left on the log-SNR distribution \ie more weight to lower log-SNR regimes. We then change the $\delta$ parameter of the sigmoid schedule to choose ($\delta=0.8, \gamma=-1$) for a peaked distribution of log-SNR around -1 and ($\delta=3.5, \gamma=0)$ for a flat distribution over noise levels.

\begin{figure}[!htb]
    \centering
    \includegraphics[width=.9\linewidth]{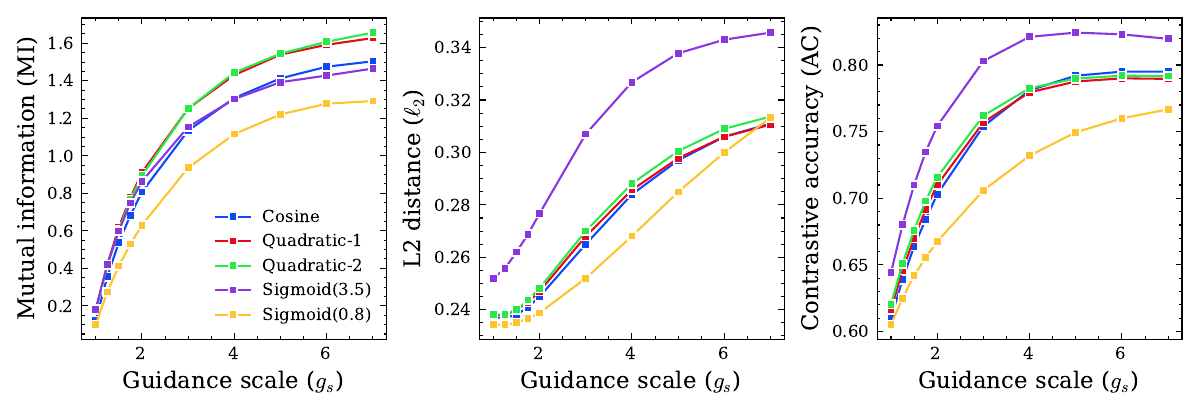}
    \caption{\textbf{Comparing noise schedules.} The prefix-suffix mutual information ($\mutinfo$), the $\ltwo$ distance and contrastive accuracy ($\mseacc$) scores of evaluated \cfour documents while varying the guidance scale $(\guidance)$ under different schedules.}
    \label{fig:arch:schedules:sweep}
\end{figure}

We follow the experimental setup described in \Cref{sec:archi:lcm:ablation:setup} and report the results of the pre-training evaluation in \Cref{tbl:archi:ablation:schedule}.
Both quadratic schedules achieve a better $\mutinfo$ score while the wide sigmoid schedule $(\delta, \gamma)=(3.5, 0)$ achieves the highest accuracy $\mseacc$ on both \cfour and \wikipedia. To further understand the differences between those schedules we re-evaluate on \cfour while varying the guidance scale $\guidance$. The results in \Cref{fig:arch:schedules:sweep} confirm that
the wide sigmoid schedule ($\delta=3.5, \gamma=0$) has a much higher contrastive accuracy across the board (rightmost panel) while closely matching the cosine schedule in terms of mutual information ($\mutinfo$).
This schedule being trained on a wider spectrum of log-SNR learns to contrast more than to regress.
Contrast this behavior to the peaked sigmoid schedule ($\delta=0.8, \gamma=-1)$ where the model focuses on regressing to the target (lower $\ltwo$), akin to a \mselcm, resulting in a lower contrastive accuracy.
\begin{table}[hbtp!]
    \centering
    \begin{tabular}{llllll|lllll}
        \toprule
        \multirow{2}*{Model} 
        & \multicolumn{5}{c|}{\cfour} 
        & \multicolumn{5}{c}{\wikipedia}  
        \\
        \cmidrule(lr){2-6} \cmidrule(lr){7-11}
         & $\ltwo$ & $\ltworound$ & $\paraphrasing$ & $\mseacc$ & $\mutinfo$ 
         & $\ltwo$ & $\ltworound$ & $\paraphrasing$ & $\mseacc$ & $\mutinfo$ 
         \\
        \midrule        
        Cosine & 0.265 & 0.261 &  2.265 & 75.4\% & 1.134
        & 0.307 & 0.297 & 2.079 & 78.8\% & 1.307 \\
        \midrule
        Quadratic-1 & 0.268 & 0.264  & 2.341  & 75.7\% & \bf 1.252
        & 0.309 & 0.300 & 2.202 & 79.1\% & \bf 1.409 \\
        Quadratic-2 & 0.270 & 0.265 & 2.320 & 76.2\% & \bf 1.252
        & 0.312 & 0.303 & 2.185  & 79.7\% & 1.399 \\ 
        \midrule
        Sigmoid(1.5, -1) & 0.257 & 0.259 & 2.226 & 74\% & 1.083
        & 0.298 & 0.292 & 2.110 & 77\% & 1.271 \\
        Sigmoid(1.5, -2) & 0.277 & 0.267 & 2.291 & 77.2\% & 1.173 &
        0.321 & 0.303 & 2.179 & 80.3\% & 1.308 \\
        Sigmoid(0.8, -1) & 0.252 & 0.255  & 2.053 & 70.6\% & 0.936 
        & 0.285 & 0.283 & 1.883 & 71.7\% & 1.127 \\
        Sigmoid(3.5, 0) & 0.307 & 0.265 & 2.347 & \bf 80.3\% & 1.154
        & 0.347 & 0.303 & 2.187 & \bf 83.7\% & 1.288 \\ 
        \toprule
    \end{tabular}
    \caption{\textbf{Comparing noise schedules.} Results of the pre-training evaluation on two corpora, \cfour and \wikipedia.}\label{tbl:archi:ablation:schedule}
\end{table}

\subsubsection{Studying the loss weighting strategies}
In this section we compare the baseline \twotower diffusion \lcm trained with the simplified objective (\ie $\omega(t)=1,\;\forall t)$) to models trained with the clamped-SNR weighting strategy. We consider two sets of $(\lambda_{\min}, \lambda_{\max})$: $(\lambda_{\min}, \lambda_{\max}) = (0, 10)$ and $(\lambda_{\min}, \lambda_{\max})=(0.001, 5)$. All models in this section are trained with the cosine noise schedule.

\begin{figure}[!htb]
    \centering
    \includegraphics[width=.6\linewidth]{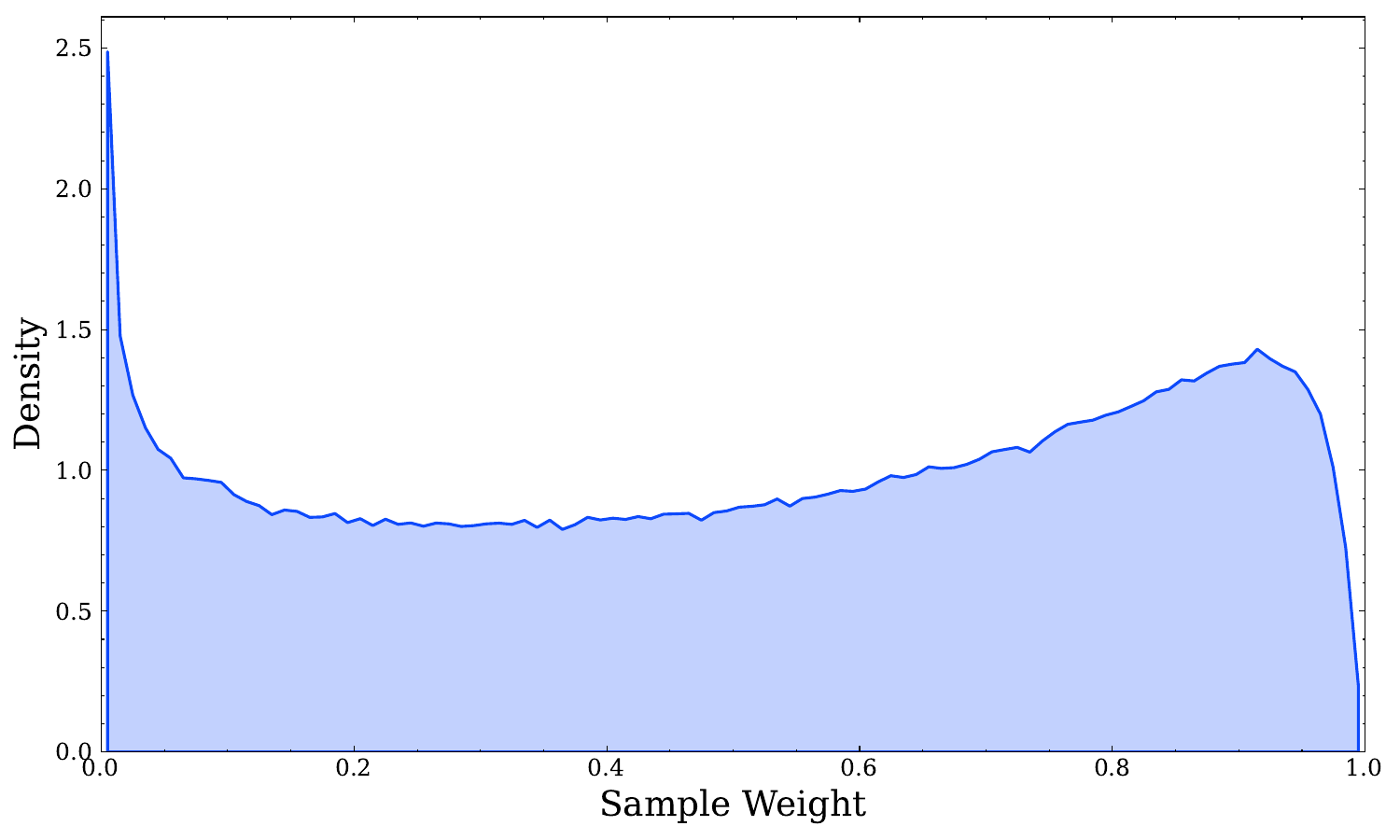}
    \caption{Resulting distribution of fragility sample weights $\omega(\diffx{0}{})$ with $\omega(\diffx{0}{}) = \sigmoid(-4 ~\rF(\rvx) + 3.5)$. 
    }\label{fig:fragility_analysis_sample_weights}
\end{figure}

\paragraph{Fragility as a sample weighing strategy}
As introduced in \Cref{eq:arch:loss:fragility}, we train in this section a \twotower \lcm model with loss terms weighted by the sample's fragility.
\begin{align}
\omega(\diffx{0}{}) &= \sigmoid(a ~\fragility(\diffx{0}{}) + b)
\end{align}
Given that estimating the fragility score of each sample as defined in \Cref{sec:analysis:fragility} can be very costly, we resort to training a simple MLP (3-layers) on 50M sampled sentences to approximate these fragility scores. This model is referred to as $\rF$. Henceforth, the sample weight is:
\begin{align}
\omega(\diffx{0}{}) &= \sigmoid(a ~\rF(\rvx) + b)
\end{align}
The two hyper-parameters ($a, b)$ are chosen so that extremely fragile sentences contribute less to the loss ($\omega(\diffx{0}{})\approx 0$),
and so that sample weights should increase smoothly as sample robustness improves. \Cref{fig:fragility_analysis_sample_weights} plots the sample weights distribution evaluated on a pre-training 
dataset with $a=-4$ and $b=3.5$.

We follow the experimental setup described in \Cref{sec:archi:lcm:ablation:setup} and report the results of the pre-training evaluation in \Cref{tbl:archi:ablation:weighting}. We observe that weighting with clamped SNRs does not improve the quality of generated texts as measured with our pre-training evaluation metrics.
When it comes to the fragility-aware weighting strategy, we observe an improvement in the contrastive accuracy of the model.
In the remainder of this work we default to the simplified training objective ($\omega(t)=1,\,\forall t)$.
\begin{table}[hbtp!]
    \centering
    \small
    \begin{tabular}{llllll|lllll}
        \toprule
        \multirow{2}*{Model} 
        & \multicolumn{5}{c|}{\cfour} 
        & \multicolumn{5}{c}{\wikipedia}  
        \\
        \cmidrule(lr){2-6} \cmidrule(lr){7-11}
         & $\ltwo$ & $\ltworound$ & $\paraphrasing$ & $\mseacc$ & $\mutinfo$ 
         & $\ltwo$ & $\ltworound$ & $\paraphrasing$ & $\mseacc$ & $\mutinfo$ 
         \\
        \midrule        
        Baseline $\omega(t)=1$ & 0.265 & 0.261 &  2.265 & 75.4\% & \bf 1.134
        & 0.307 & 0.297 & 2.079 & 78.8\% & \bf 1.307 \\
        \midrule
        SNR (0,10) & 0.280 & 0.264 & 2.334 & 74.8\% & 1.107
        & 0.320 & 0.296 & 2.102 & 77.9\% & 1.212\\
        SNR (0.001,5) & 0.266 & 0.261 & 2.269 & 73.4\% & 1.094 & 
        0.304 & 0.291 & 2.007 & 76.6\% & 1.295\\
        \midrule
        Fragility & 0.2815 & 0.273 & 2.306 & \bf 76.5\% & 1.103  & 0.321 & 0.308 & 2.118 & \bf 80.7\% & 1.193 \\
        \toprule
    \end{tabular}
    \caption{\textbf{Comparing weighting strategies.} Results of the pre-training evaluation on two corpora, \cfour and \wikipedia.}\label{tbl:archi:ablation:weighting}
\end{table}

\subsection{Analysis}
\label{sec:analysis}

\subsubsection{Inference efficiency of \lcms}
\begin{figure}
    \centering
    \includegraphics[width=1\linewidth]{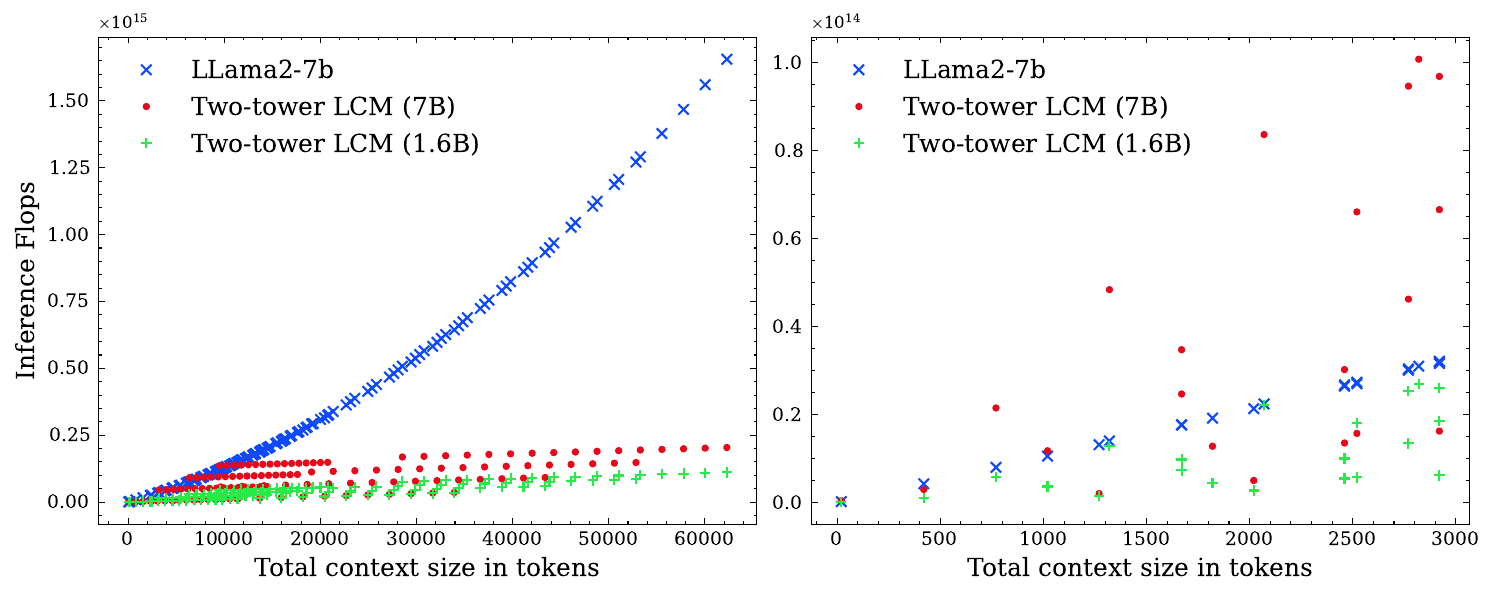}
    \caption{\textbf{Theoretical inference Flops of \lcms and LLLms}. We evaluate the inference flops for different text lengths (in \llamatwo tokens) with a variable average sentence length. Only extremely short sentences ($\le 10$ tokens) favor LLMs.}
    \label{fig:ablation:flops}
\end{figure}

We compare in this section the inference computational cost of the \twotower \lcm to that of a vanilla LLM as a function of the total length in tokens of the prompt and output combined. We chose the theoretical number of FLOPs, independent of any specific optimization. These optimizations are generic to the transformer architecture and also apply to our \lcm. 
We include in this comparison two configurations of \lcms; the 1.6B used in the previous ablation studies and a 7B model we scale up to in the following sections. For both \lcms, we estimate the inference cost with inference sample steps $\rS=40$.
Given the quadratic complexity of the attention mechanism in transformers, the complexity sharply increases with the context size (see upper right corner of \Cref{fig:ablation:flops}'s left panel). 
The complexity of the \lcm depends on how the context is sentencized: a context length of 200 tokens split into 10 sentences (20 tokens each) will incur a higher cost than the same 200 tokens split into 5 sentences (40 tokens each). We account for this by computing the cost on a range of sentence lengths but report the total context size on the x-axis (context size = sentence length $\times$ number of sentences).
The \lcm  shows substantially better scalability with respect to increasing context size. The inference computational cost of the \lcm includes the three steps of (1) encoding into \sonar, (2) \lcm  prediction in the sentence space then (3) decoding with a \sonar decoder. The inference cost of \lcms varies significantly depending on the average length in tokens per sentence. 
For extremely short sentences (less than 10 tokens), an \llm is more computationally efficient (see lower left corner of \Cref{fig:ablation:flops}'s right panel). 

\subsubsection{Fragility of \sonar space}
\label{sec:analysis:fragility}

When we perform modeling in a latent space, we primarily rely on the induced geometry ($L2$-distance).
However, the homogeneous Euclidean geometry of any latent representation will not perfectly match the underlying text semantics. %
This is evidenced by the fact that 
a small perturbation in the embedding space may result in a drastic loss of semantic information after decoding.
We dub such embeddings \emph{``fragile''}.
For this reason, we aim to quantify the fragility of semantic embeddings (namely \sonar codes) to understand the quality of the \lcm training data and how this fragility can hinder the \lcm training dynamics.

Given a text fragment $w$ and its \sonar code $\rvx = \encode(w)$, we define the fragility of $w$ as:
\begin{align}
\fragility(w) & \defeq - \ds \mathbb{E}_{\alpha \sim \uniform([0, 1]),\ \rvepsilon \sim \gaussian(\rvzero, \rmI)} \left[ \simil(w, w_{\alpha, \rvepsilon}) \right],\label{eqn:fragility:score}\\
\rvx_{\alpha, \rvepsilon} & = \denormalize\left(\sqrt{1 - \alpha} ~ \normalize(\rvx) + \sqrt{\alpha} ~ \rvepsilon\right),\label{eqn:fragility:noised-sample}\\
w_{\alpha, \rvepsilon} & = \decode(\rvx_{\alpha, \rvepsilon}),
\end{align}

where $\normalize$ and $\denormalize$ are the normalization and denormalization operators introduced in \Cref{eq:sonar:normalizer} with the goal of making $\rvx$'s coordinates scale-independent.
The ``$\encode$`` operation maps text fragments into \sonar space, and the ``$\decode$`` operation produces a text fragment from a given vector in the \sonar space.
For each $\alpha$ in $[0, 1]$, $\rvx_{\alpha, \rvepsilon}$ is the perturbed version of $\rvx$ where a noise vector of variance $\alpha$ is linearly combined with $\rvx$. The perturbed vector is then decoded into a text fragment $w_{\alpha, \rvepsilon}$.
This perturbation is similar to the variance-preserving noising used in diffusion \lcms (see \Cref{sec:arch:diffsion:intro}).

The ``$\simil$'' operator in \Cref{eqn:fragility:score} is set to be a semantic similarity metric comparing the perturbed text $w_{\alpha, \rvepsilon}$ to the original $w$. We considered the following options:
\begin{itemize}[style=unboxed,leftmargin=*]
    \item \textbf{Auto-Encoding \bleu.} $\simil(w, w_{\alpha, \rvepsilon}) = \bleu(w, w_{\alpha, \rvepsilon})$.
    \item \textbf{External cosine similarity.} Provided an external text encoder (read unrelated to \sonar) that encodes a text fragment $w$ into $\externalencode(w)$
        $\simil(w, w_{\alpha, \rvepsilon}) = \cossim(\externalencode(w), \externalencode(w_{\alpha, \epsilon}))$, where $\cossim$ is the cosine similarity measure.
        Compared to Auto-Encoding \bleu, this method is typically more robust to paraphrasing.
\end{itemize}
\paragraph{Finetuned robust decoder.}
To serve as a testbed for our fragility analysis, a new \sonar decoder for English text is finetuned on a sample of our pre-training data. In order to improve the decoder's robustness to imperfectly generated embeddings from the \lcm, we follow \Cref{eqn:fragility:noised-sample} and add random noise vectors to \sonar embeddings during training.
As reported in \Cref{tab:new_noisy_decoder}, the finetuned \sonar decoder exhibits stronger performance across a range of public corpora.

\begin{table}[!htb]
    \centering
    \begin{tabular}{lcccc}
        \toprule
        Model & $\textsc{Flores}$ & \cnndm & $\gutenberg$ & $\cfour$ \\
        \midrule
        \textsc{Base} \sonar decoder &  79.5 & 75.9 & 70.5 & 75.7 \\
        \textsc{Finetuned} \sonar decoder & 88.0 & 87.6 & 85.6 & 87.5 \\
        \bottomrule
    \end{tabular}
    \caption{\textbf{Comparing \sonar decoders.} Raw reconstruction performance of our base \sonar decoder vs. the new decoder trained with noised embeddings. Scores are Auto-Encoding BLEU on random subsets of 10k sentences from each dataset, except for \flores where we use the entire dev split.}
    \label{tab:new_noisy_decoder}
\end{table}

\paragraph{Fragility study.}
We sample 50M random text fragments, and for each sample we generate 9 perturbations corresponding to different noise levels $\alpha \in [0.1, 0.2, \ldots, 0.9]$.
For the external cosine similarity metric we use \textsc{mGTE} as external encoder~\citep{zhang2024mgtegeneralizedlongcontexttext}.

\begin{figure}[!htb]
    \centering
    \includegraphics[width=.95\linewidth]{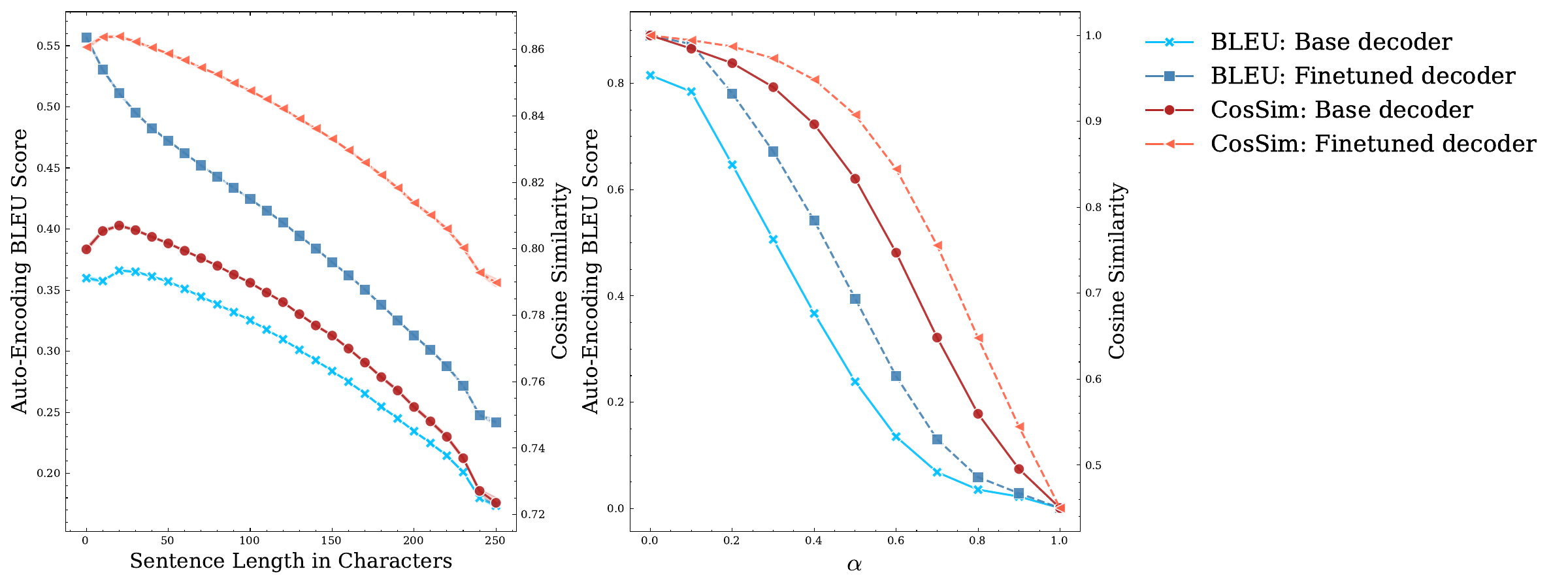}
    \caption{\textbf{Fragility scores.} Auto-Encoding BLEU and external cosine similarity. In the left-hand panel as a function of the text length ($\alpha$-averaged)  and in the right-hand panel as a function of the noise variance $\alpha$.
}
\label{fig:fragility_analysis}
\end{figure}

\begin{figure}[!htb]
    \centering
    \includegraphics[width=.65\linewidth]{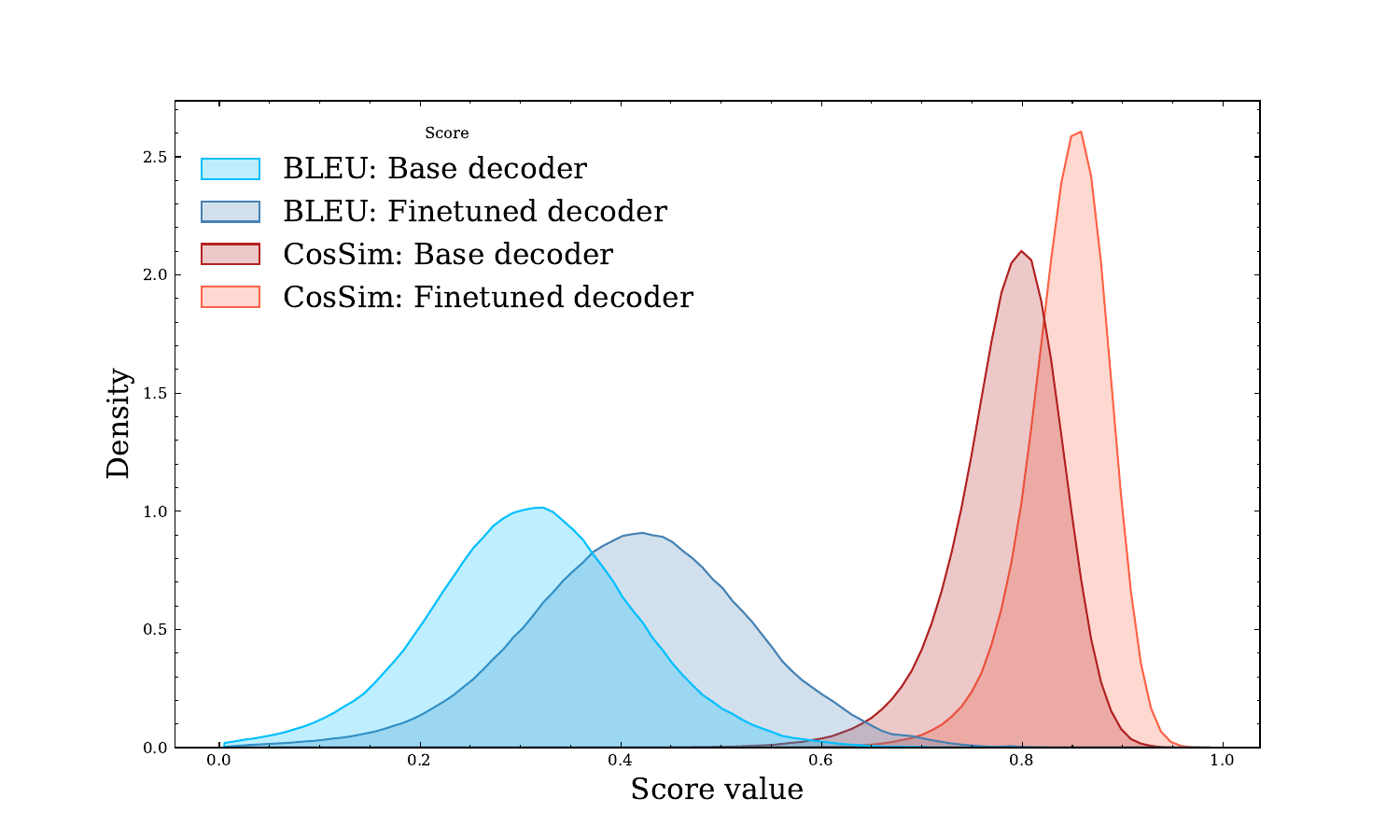}
    \caption{Auto-Encoding BLEU scores and cosine similarity ($\alpha$-averaged) distributions.}
    \label{fig:fragility_analysis_dist}
\end{figure}

We depict in the right panel of \Cref{fig:fragility_analysis} the curves of both score functions with respect to the noise level $\alpha$.
We observe that $\bleu$ scores decrease faster than the cosine similarity.
Most importantly, fragility scores are sensitive to the choice of the decoder. In particular, 
both Auto-Encoding BLEU and cosine similarity scores decrease at a markedly slower rate for the \textsc{Finetuned} decoder than for the \textsc{Base} one as the amount of noise increases.
 We note also that the overall score  distribution (after averaging over all $\alpha$), shown in \Cref{fig:fragility_analysis_dist}, exhibits a large spread of fragility scores across \sonar samples.

One factor that can explain such a discrepancy is the text length. Compared to the Auto-Encoding BLEU metric (which drops only by 1--2\% for long sentences),
fragility is more sensitive to the length of sentences and drops faster for both similarity metrics.
This shows that using a max sentence length over $250$ can be extremely challenging for \sonar and the \lcm model.
On the other hand, even if short sentences are on average more robust, 
splitting a long sentence in the wrong place may result in shorter but more fragile sub-sentences.

Taking a closer look at the 5\% most fragile embeddings, we notice that they are very noisy. 
Typically, they correspond to hyperlinks, references, unique ids, code-switched or numerical entries. These are likely artifacts that the \sonar models were not exposed to during training or where the \sonar tokenizer fails.
Fragility can thus be used to filter out hard samples from the training data.
We also observe that short but complex technical phrases can be more fragile than common language phrases of similar length.

\section{Scaling the model to 7B}
\label{sec:bigmodel}

This section describes our effort to scale our model to 7B parameters and compare the performance against other approaches such as token-based LLMs on more challenging tasks such as summarization and summarization expansion (detailed in \Cref{sec:task}).

Based on the results in \Cref{sec:archi:lcm:ablations} where the two diffusion-based \lcms (\interleaved and \twotower) outperform the other variants, we decided to scale a diffusion model to 7B parameters.
We chose to scale \twotower given its smaller memory footprint, particularly when processing long contexts with a shallower \ctxenc tower 

The large 7B \twotower diffusion \lcm has 5 layers in its \ctxenc and 14 layers in its \denoiser.
Its dimension has been extended to $\modeldim{=}4096$.
Each self-attention layer has 32 attention heads.
All other parameters are kept the same as for the 1.6B \twotower model. 
The model is pre-trained on a dataset of 2.3B documents, representing 2.7T tokens and 142.4B concepts/sentences.
We pre-trained this model on Meta’s RSC for 124k optimization steps spanning 256 A100 GPUs with a total batch size of 1M concepts. We further extend the context length of this model to cover 2048 concepts instead of the 128 concepts in the ablation experiments.
We trained using the AdamW optimizer with $(\beta_1, \beta_2)=(0.9, 0.95)$, $\epsilon=1e$-$5$ and a weight decay of 0.1. We use a cosine learning rate schedule, with warm-up of 10,000 steps up to $\text{LR}=3e$-$4$. To improve training stability we clip gradients at a maximum norm of $\rg=10$.
We subsequently finetune the 7B \twotower \lcm on publicly available instruction tuning datasets following~\citet{chung2024scaling}. 
Each sample consists of a prompt and an answer and we back-propagate on answer sentences only. Each answer sequence is suffixed with the phrase ``End of response.'' to teach the model when to stop generating.
The finetuning data totals 
389M sentences, of which 53M are answers 
(\ie targets).
For supervised finetuning, we use a cosine learning rate schedule with an initial rate of $\text{LR}=3e$-$5$ and finetune the model for 7 epochs with a batch size of 
262K sentences (prompts and answers combined).
We will refer to the pre-trained model as \bigtwotower and the finetuned model as \IFTtwotower. 
\subsection{Evaluation Tasks and Data}
\label{sec:task}

This section describes the tasks on which we are evaluating and benchmarking our proposed model. We detail datasets, baselines and metrics. 
For each task, the dataset was processed with the same sentence splitter and \sonar encoder as used in the \lcm training.

\subsubsection{Metrics}
\label{sub:summ_metrics}

As longform text generation is the main challenge for \lcm, our benchmarking is mainly focused on generative tasks, which are notoriously more difficult to evaluate automatically. Therefore, we evaluate them with multiple automatic metrics, chosen to focus on complementary aspects on generation quality. All metrics used in this section are summarized in \Cref{tab:metrics}. 

For summarization and summary expansion (defined below), we report the traditional reference-based \rougellong metric \citep{lin2004rouge}. As summarization models have a tendency to copy content from the source or from its own generated prefix, we report two additional word-based metrics. To evaluate how much content is directly copied from the source, we report the proportion of word 3-grams of the source that are present in the output (\ovlthree). To evaluate repetitiveness of the generated texts, we report the portion of duplicated word 4-grams in the output (\repfour). 

To complement word-based metrics with summarization-focused neural evaluation, we use two metrics introduced by \citet{clark2023seahorse}: average probabilities of the SEAHORSE classifiers for Q4 (whether all the information in the summary is fully attributable to the source), denoted as \shfour in the following and Q5 (whether the summary captures the main ideas of the source), denoted as \shfive.

As a metric of the overall fluency of the generated sentences, we report an average probability that the sentence is linguistically acceptable, as predicted by a classifier trained by \citet{sent_fluency_style20} on the \cola dataset \citep{warstadt-etal-2019-neural}, further referred to as \cola. 
To evaluate the local coherence of the generated text, we report the average cosine similarity between each $n$'th and $n+2$'th sentence \citep{parola2023speech}.

\begin{table}[h!]
\centering
\scriptsize
\begin{tabular}{p{2cm}p{2.5cm}lp{5.5cm}p{3cm}}
\toprule
{\bf Task} & {\bf Area} & {\bf Metric} & {\bf Description} & {\bf Reference} \\  \midrule
{\bf Summarization} & Target similarity & $\rougel$ & ROUGE-L & ~\citet{lin2004rouge}  \\ 
& Source similarity & \ovlthree & N-grams overlap (N=3) & \\ 
& Grammaticality & \repfour & Portion of duplicated N-grams (N=4) & \citet{welleck2019neural} \\
& Fluency & \cola & Sentence fluency classifier score & ~\citet{sent_fluency_style20} \\
& Attribution & \shfour & Seahorse-Large-Q4 score & \citet{clark2023seahorse} \\
& Semantic coverage & \shfive & Seahorse-Large-Q5 coverage score & \citet{clark2023seahorse} \\
\midrule
\MR{2}{2cm}{\bf Summary Expansion} & Grammaticality & \repfour & (see above) & ~\citet{welleck2019neural} \\
& Fluency & \cola & (see above) &  ~\citet{sent_fluency_style20} \\
\bottomrule
\end{tabular}
\caption{\label{tab:metrics}Summary of automatic metrics used in different tasks in \Cref{sec:task}. Order mostly follows paper's narrative.}
\end{table}

\subsubsection{Summarization}
\label{sub:summ}

\paragraph{Task and datasets.}
When considering a relatively long document, a summarization task can be described as the act of generating a much shorter corresponding document that includes the essential information contained in the long document and the same logical structure linking the various pieces of essential information.

Summarization techniques can range from more extractive to more abstractive. Extractive techniques attempt to preserve in the summary the same vocabulary as that found in the long document, thereby shortening the long by removing details and superfluous wording. Abstractive techniques, on the other hand, attempt to produce the summary by rephrasing the essential pieces of information found in the long document. Our work focuses more on abstractive summarization, as such type of summarization cannot be performed without some form of understanding and reasoning.

We use the \cnndm~\citep{cnndm:neurip:2015} and \xsum~\citep{xsum:2018} datasets. %
We also report results on the challenging \lcfo corpus which takes long documents as input, approx. 5k words \citep{lcfo}. The task is to provide abstractive summaries with lengths representing 20\%, 10\%, and 5\% of the input document.
Detailed statistics are provided in \Cref{tab:bigmodel:eval_data_statistics}.

\begin{table}[!htb]
\centering
\small
\begin{tabular}{@{}lrrrrrrrH@{}}
\toprule
 & & \multicolumn{3}{c}{\#\llamatwo Tokens} & \multicolumn{3}{c}{\#Sentences} & \multirow{2}*{\makecell[c]{Total\\sentences}} \\
\cmidrule(lr){3-5}  \cmidrule(lr){6-8} 
Dataset & \#Docs & Q1 & Q2 & Q3 & Q1 & Q2 & Q3 &  \\
\midrule

\cnndm & 11.5k & 605/61 & 892/78 &	1266/97  &
10/3 & 14/4	& 21/4 & 189k/45k\\
\xsum  & 11.3k & 273/25   &	445/30   &  735/35   
             & 25/1  & 30/1  & 35/1  & 193k/11k \\

\lcfofive & 249 & 6559/341 & 7214/378  & 7916/418	 & 209/12 & 295/15 & 527/18 & 99k/4k  \\
\lcfoten & 249 & 6559/654  & 7214/718  & 7916/796	 & 209/22 & 295/27 & 527/32	& 99k/7k  \\
\lcfotwenty &  249 &  6559/1276 &  7214/1403 & 7916/1524	 & 209/41  & 295/48  &  527/59	& 99k/13k \\
\bottomrule
\end{tabular}
\caption{Statistics of the test split of evaluation benchmarks. For each subset we report the number of documents and statistics of document and summary length in terms of sentences and \llamatwo tokens. Each table cell shows ``document/summary'' length quartiles.}
\label{tab:bigmodel:eval_data_statistics}
\end{table}

\paragraph{Baselines.} For \cnndm and \xsum, we compare against several baselines of different architectures (encoder-decoder transformer, decoder-only LLMs) that are known to perform well on summarization tasks. For encoder-decoder transformer models, we use T5 \citep{T5:2020:jmlr}.
For decoder-only LLMs, we choose \gemma-7B, \llama-3.1-8B and \mistral-7B-v0.3. We chose the published instruction-tuned models to compare with the \lcm with the same training regime, and have a similar size (7B). Note that while T5 has much smaller sizes than the \lcm, this is compensated by using models that are fine-tuned explicitly on the target evaluation dataset.

\begin{table*}[t!]
\centering
\small
\resizebox{\textwidth}{!}{%
\begin{tabular}{@{}lp{1.5cm}rcccccccc@{}}
\toprule
\textsc{Model} & Paradigm & \multicolumn{8}{c}{\bf\cnndm} \\
\midrule
& & $\rougel$($\uparrow$) & \ovlthree($\uparrow$) & \repfour($\downarrow$) & \cola($\uparrow$) & \shfour($\uparrow$) & \shfive($\uparrow$) \\
\cmidrule{3-10}
Ground truth & --- & 100.00 & 0.170 & 0.684 & 0.850 & 0.683 &  0.586  \\
\cmidrule{0-10}

T5-3B & SFT & \bf 37.56 & 0.174 & 0.854 & 0.946 & 0.773 & 0.503   \\

\cmidrule{0-10}
\gemmaIT & IFT & 31.14 & 0.245 & 1.032 & 0.963 & 0.740 & 0.560  \\

\mistralIT & IFT & 36.06 & 0.200 & 0.780 & 0.972 & \textbf{0.780} & 0.676   \\

\llamaIT & IFT & 34.97 & \textbf{0.248} & 0.928 & \textbf{0.973} & 0.763 & \textbf{0.692}   \\
\addlinespace[0.5em]
\cmidrule{0-10}

\IFTtwotower & IFT & \textbf{36.47} & 0.177 & \textbf{0.757} & 0.767 & 0.723 & 0.459  \\

\bottomrule
\toprule
\textsc{Model} & Paradigm & \multicolumn{8}{c}{\bf \xsum} \\
\midrule

& & $\rougel$($\uparrow$) & \ovlthree($\uparrow$) & \repfour($\downarrow$) & \cola($\uparrow$) & \shfour($\uparrow$) & \shfive($\uparrow$) \\
\cmidrule{3-10}

Ground truth & --- & 100.00 & 0.108 & 0.399 & 0.987 & 0.352 & 0.418  \\
\cmidrule{0-10}
T5-3B & --- & 17.11 & 0.221 & 0.671 & 0.939 & 0.680 & 0.450 \\

\cmidrule{0-10}
\gemmaIT & IFT & 18.20 & 0.177 & 0.620 & 0.769 & 0.546 & 0.446 \\
\mistralIT & IFT & 21.22 & 0.162 & 0.480 & 0.922 & 0.633 & 0.621   \\
\llamaIT & IFT & 20.35 & \textbf{0.186} & 0.501 & \textbf{0.941} & \textbf{0.687} & \textbf{0.658}  \\
\cmidrule{0-10}

\IFTtwotower & IFT & \textbf{23.71} & 0.106 & \textbf{0.464} & 0.683 & 0.358 & 0.284  \\

\bottomrule
\end{tabular}%
}%
\caption{Performance on the \cnndm and \xsum summarization tasks.}
\label{tab:summarization:daily}
\end{table*}

\paragraph{Summarization results.}
\Cref{tab:summarization:daily} contains the results of different baselines and our \lcm model for summarization (\cnndm and \xsum).
We can notice that the \lcm produces competitive \rougellong scores when compared to a specifically tuned \llm (T5-3B) and even surpasses the instruct-finetuned \llms. 
Our model tends to generate more abstractive summaries rather than extractive ones, as shown by the lower \ovlthree scores. 
The \lcm produces fewer repetitions compared to \llms, and more importantly, the repetition rate is closer to the ground truth one. 
The \lcm generates globally less fluent summaries according to \cola classifier. 
However, we can remark that even the human generated ground truth gets a lower score compared to the \llm. 
A similar behavior is observed for the source attribution (\shfour) and semantic coverage (\shfive). 
This may be explained by model-based metrics that are more biased towards \llm generated content.

\paragraph{Long-context summarization results.}
\Cref{tabl:summ-long} presents the results for long-context summarization (\lcfofive, \lcfoten and \lcfotwenty).
This is a challenging task for most of the models.
For example, \mistralIT seems to be unable to follow the length instruction of the summary--it always generates summaries which length is about 50\% of the source. \mistralIT also has the highest \shfour score, \ie source attribution.
The summaries generated by \gemmaIT tend to be longer than requested, while \llamaIT generates summaries which length is the closest to the requested size.

The \lcm has only seen a limited amount of long documents in the pretraining and fine-tuning data. Nevertheless, it performs well for this task.
It outperforms \mistralIT and \gemmaIT in the metric $\rougellong$ for the 5 and 10\% conditions, and is close to \gemmaIT for the 20\% condition. We also observe that the \lcm yields high \shfive scores for all conditions, \ie the summaries can be attributed to the source.

\begin{table*}[t!]
\centering
\small
\resizebox{\textwidth}{!}{%
\begin{tabular}{@{}lp{1.2cm}rcccccccc@{}}
\toprule
\textsc{Method} & WR & \multicolumn{8}{c}{\lcfofive} \\
\midrule

& & $\rougel$($\uparrow$) & \ovlthree($\uparrow$) & \repfour($\downarrow$) & \cola($\uparrow$) & \shfour($\uparrow$) & \shfive($\uparrow$) \\
\cmidrule{3-10}
\gemmaIT & 0.107 & 25.21 & 0.151 & 4.711 & 0.688 & 0.357 & 0.174 \\

\mistralIT & 0.512 & 21.36 & \textbf{0.532} & 5.997 & 0.854 & \textbf{0.656} & 0.296\\

\llamaIT & 0.076 & \textbf{37.67} & 0.190 & 2.767 & \textbf{0.931} & 0.488 & \textbf{0.314} \\
\midrule

\IFTtwotower & 0.060 & 26.88 & 0.162 & \textbf{2.473} & 0.796 & 0.628 & 0.196 \\
\bottomrule
& & \multicolumn{8}{c}{\lcfoten} \\
& & $\rougel$($\uparrow$) & \ovlthree($\uparrow$) & \repfour($\downarrow$) & \cola($\uparrow$) & \shfour($\uparrow$) & \shfive($\uparrow$) \\
\cmidrule{3-10}
\gemmaIT & 0.150 & 29.25 & 0.164 & 6.427 & 0.667 & 0.377 & 0.194 \\

\mistralIT & 0.549 & 25.00 & \textbf{0.537} & 6.289 & 0.848 & \textbf{0.660} & 0.306 \\

\llamaIT & 0.128 & \textbf{42.85} & 0.243 & 3.804 & \textbf{0.907} & 0.486 & \textbf{0.310} \\
\cmidrule{0-10}

\IFTtwotower & 0.089 & 29.38 & 0.202 & \textbf{3.00} & 0.791 & 0.623 & 0.183 \\

\bottomrule
& & \multicolumn{8}{c}{\lcfotwenty} \\
& & $\rougel$($\uparrow$) & \ovlthree($\uparrow$) & \repfour($\downarrow$) & \cola($\uparrow$) & \shfour($\uparrow$) & \shfive($\uparrow$) \\
\cmidrule{3-10}
\gemmaIT & 0.257 & 33.32 & 0.201 & 9.188 & 0.603 & 0.425 & 0.239 \\

\mistralIT & 0.493 & 28.82 & \textbf{0.527} & 5.806 & 0.858 & \textbf{0.658} & 0.293 \\

\llamaIT & 0.179 & \textbf{46.92} & 0.272 & 4.783 & \textbf{0.888} & 0.485 & \textbf{0.315} \\
\cmidrule{0-10}

\IFTtwotower & 0.140 & 31.74 & 0.253 & \textbf{3.664} & 0.779 & 0.613 & 0.187 \\

\bottomrule
\end{tabular}%
}
\caption{\label{tabl:summ-long}Performance on the long-context summarization task of LCFO. WR is the word count ratio between the generated text and the source document.
}
\end{table*}Finally, we observe that \llamaIT performs substantially better than the other \llms, according to $\rougellong$, while all \llms have similar performance on the \cnndm and \xsum summarization tasks. This could be explained by training data contamination for \llamaIT, or by the fact that the other two \llms struggle to handle the long input context.\FloatBarrier
\section{\LCM Extensions}

In this section, we explore several extension of the \LCM. First, we evaluate the \lcm on the new task of summary expansion, \ie given a summary, create a longer text. We then showcase the good zero-shot generalization performance of the \lcm. Finally, we explore an approach to add higher level information beyond sentences.

\subsection{\summaryexpansion}
\label{sub:summexp}

\paragraph{Task and datasets.}
When considering a short and concise document that has similar properties to those of a summary (i.e., mainly a stand-alone document that abstracts from details), a summary expansion task can be described as the act of generating a much longer document that preserves the essential elements found in the corresponding short document, as well as the logical structure that connects such elements. As this is a more freely generative task, an additional requirement to be taken into consideration is that of coherence (for example, the detailed information included in one generated sentence should not contradict that included in another sentence). The summary expansion task presented here consists in taking summaries as inputs, from \cnndm and \xsum, and generating a long document.
Note that the goal is not to recreate the factual information of the initial document rather than evaluating the capability of the model to extend the input text in a meaningful and fluent way.
We use similar baselines and metrics as in the previous section \ref{sub:summ}.

\begin{table*}[!ht]
\centering
\small
\begin{tabular}{@{}lrrcccc@{}}
\toprule

\multicolumn{6}{c}{\cnndm} \\
\midrule
\textsc{Method} & WR & $\rougel$($\uparrow$)
& \ovlthree($\uparrow$)
& \repfour($\downarrow$)
& \cola($\uparrow$) \\
\cmidrule(lr){2-6}
\gemmaIT & 6.8 & 35.54 & 0.801 & 2.104 & 0.951 \\

\mistralIT & 6.4 & 34.24 & 0.817 & \textbf{2.063} & \textbf{0.959} \\

\llamaIT & 8.5 & \textbf{37.76} & \textbf{0.822} & 2.582 & 0.844 \\
\cmidrule{0-6}
\IFTtwotower & 6.3 & 30.85 & 0.726 & 2.911 & 0.474 \\
\bottomrule
\toprule
\multicolumn{6}{c}{\xsum} \\
\midrule
\textsc{Method} & WR & $\rougel$($\uparrow$)
& \ovlthree($\uparrow$)
& \repfour($\downarrow$)
& \cola($\uparrow$) \\
\cmidrule(lr){2-6}
\gemmaIT & 19.5 & 17.89 & \textbf{0.963} & 10.238 & 0.116 \\

\mistralIT & 1.6 & \textbf{29.31} & 0.893 & 2.268 & \textbf{0.939} \\

\llamaIT & 19.8 & 28.84 & 0.915 & 2.543 & 0.898 \\
\cmidrule{0-6}
\IFTtwotower & 7.1 & 23.82 & 0.561 & \textbf{1.542} & 0.603 \\
\bottomrule
\end{tabular}%
\caption{Performance on the summary expansion tasks of \cnndm and \xsum, evaluated with the metrics described in \Cref{tab:metrics}. WR is the word count ratio between the hypothesis and the source summary.}
\label{tabl:i-summ}
\end{table*}

\paragraph{Results.}
\Cref{tabl:i-summ} shows the results of the summary expansion for \cnndm and \xsum.
First of all, regarding the word count ratio, we can see different behaviours for the two corpora.
For \cnndm, the models tend to generate texts that are 6 times larger than the input. \llamaIT produces even longer outputs (factor 8 instead of 6).
But for \xsum, while \gemmaIT and \llamaIT generate very long texts (almost 20 times longer than the prompt), the \lcm generates output of approximately the same length ratio as for \cnndm. Only \mistralIT fails to generate long outputs for this corpus.

Then, we clearly see a different trend compared to the summarization task. 
The \llms get higher \rougellong scores compared to the \lcm. 
As mentioned above, the goal of this task is not to recreate the original document. However, the $\rougel$ score tells us how much of the content of the full document can be recreated.
Contrary to the \llms, our model tends to generate different sentences compared to the original document from which the summary has been created (with an exception for \gemmaIT on \xsum). 
This is expected since our model generates embeddings that are then processed by a decoder trained on a translation task that tends to paraphrase the initial content.
However, the \cola results show that this comes along with lower fluency, especially for \cnndm.

\subsection{Zero-shot generalization performance}

\sonar is a semantic space that can represent 200 languages.
In this paper, all experiments presented so far have been done on English text.
In this section, we explore the capability of our proposed \lcm approach to process other languages in a zero-shot fashion by leveraging \sonar's ability to represent multilingual data.

We use  the \xlsum~\citep{hasan-etal-2021-xl} corpus, a large scale multilingual abstractive news summarization benchmark covering 45 languages. We score model outputs using the multilingual rouge scoring scripts released with the benchmark.\footnote{\url{https://github.com/csebuetnlp/xl-sum/tree/master/multilingual_rouge_scoring}} Note that \rougellong scores are heavily dependent on language-specific text tokenization and stemming. Unless provided in the aforementioned scripts, we tokenize the model outputs and references with the default tokenization from \citet{lin2004rouge}. Languages like Korean, Telugu and Tamil can benefit from a more appropriate stemming and tokenization.

We compare the LCM performance with \llamaIT which officially supports eight languages: English, German, French, Italian, Portuguese, Hindi, Spanish, and Thai. According to \citet{llama3.arxiv.2024}, the model has seen many additional languages during pretraining, but was instruction finetuned on those eight languages only. The \lcm, on the other hand, has never seen any language other than English, and we do not use the English \xlsum training data either.

\begin{figure}[!t]
    \centering
    \includegraphics[width=1\linewidth]{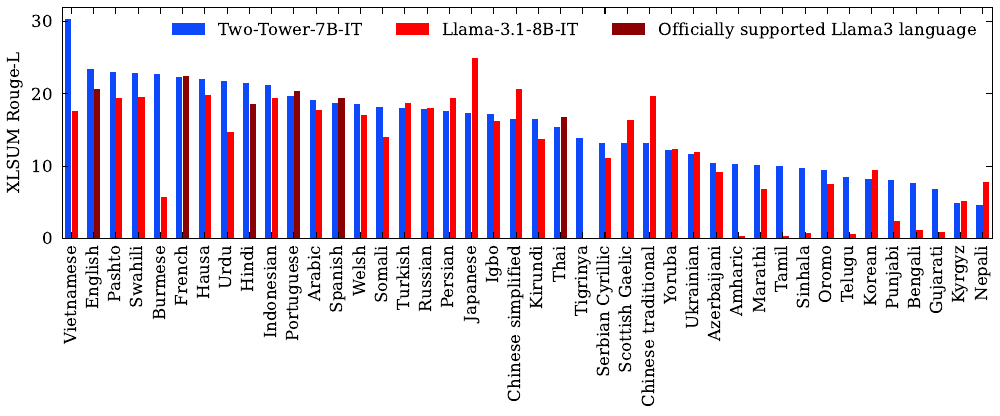}
    \caption{$\rougellong$ scores on \xlsum for \llamaIT and \IFTtwotower.}
    \label{fig:scale:xlsum}
\end{figure}

We report \rougellong scores for 42 languages in \Cref{fig:scale:xlsum}. Three languages were excluded since they are currently not supported by \sonar: Pidgin, Serbian in Latin script and Uzbek in Cyrillic script.
The \lcm substantially outperforms \llamaIT on English (23.5 compared to 20.7 \rougellong) and on the average over all six languages officially supported by both models and included in \xlsum (20.2 versus 19.7 \rougellong).\footnote{English, French, Hindi, Portuguese, Spanish and Thai.}
We also observe that the \lcm generalizes very well to many other languages, in particular low-resource languages like Southern Pashto, Burmese, Hausa or Welsch which all have $\rougellong$ scores greater than 20. Other well performing low-resource languages are Somali, Igbo or Kirundi. Finally, the \lcm obtains a $\rougellong$ score of 30.4 on Vietnamese.
Overall, these results highlight the impressive zero-shot generalization performance of the \lcm to languages it has never seen. \subsection{Exploring explicit planning}
\label{sec:planlcm}

When writing long-form text, it is often practical to first think about how to structure our narrative. Indeed \citet{vandijk} states that all texts have an innate macrostructure \ie a global discourse structure spanning the entirety of the text. In general scientific discourse, one such macrostructure is that of problem-solving \citep{heffernan-teufel-2022-problem}, where an author must first motivate and describe the problem, before proposing an appropriate solution.

Composing such a macrostructure is no easy task. Yet in the realm of \llms, there have been many recent efforts to help guide model generation using similar structures. One such popular approach is that of creating outlines, which provide a high-level overview of the desired narrative \citep{li2024advancingpreciseoutlineconditionedtext}. An alternative approach to outlines is creating summaries of future targets, which the model can then expand upon \citep{sun-etal-2022-summarize}. 

Given the nature of the \lcm operating at the concept level, it naturally creates long-form output. Therefore, it is important to ensure that the model is capable of creating coherent generations given the multitudes of possibilities for next concept prediction. In order to address this, we envision an 
explicit capability for planning. Similar to creating summaries, we propose a complementary \emph{planning model} which creates a high-level overview of what should be generated next, given the prior context. The proposed plan could span multiple concepts, such as a paragraph. The \lcm is conditioned on this plan, before it then generates the subsequent output sequence. 

\begin{figure}[ht!]
    \centering
    \includegraphics[width=0.60\linewidth]{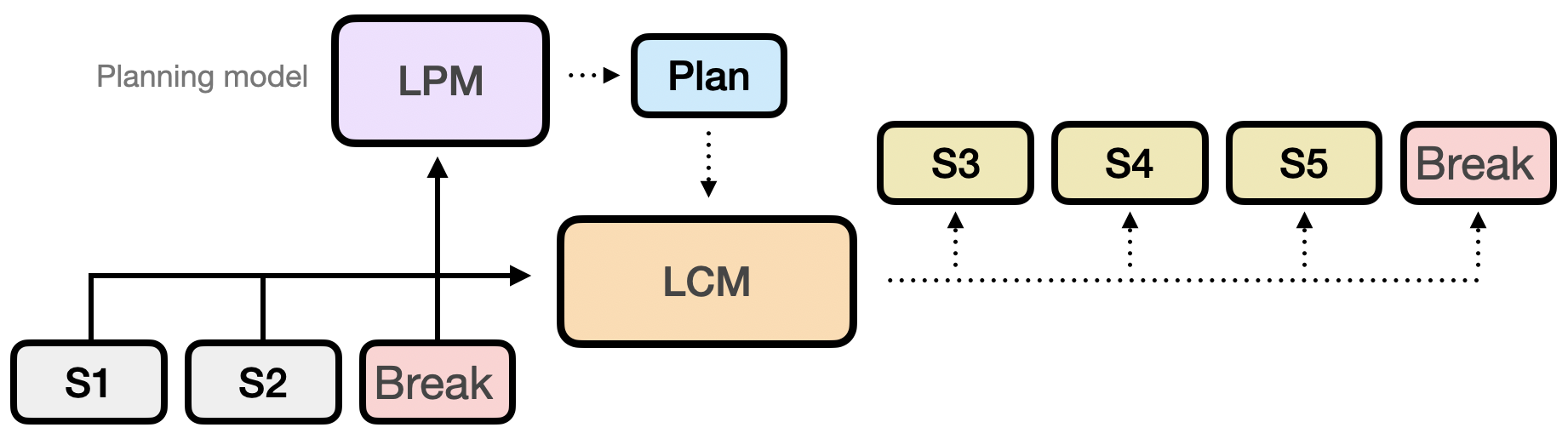}
    \caption{\lcm conditioned on both the prior context and a high-level plan for the next sequence.}
    \label{fig:three_level_lcm}
\end{figure}

Operationally the model predicts auto-regressively a sequence of concepts followed by a \emph{break} concept, which represents a natural topic cadence such as a paragraph break. Once the \emph{break} concept is predicted, the large planning model (LPM) generates a plan in order to condition the \lcm for prediction of the subsequent sequence. The model then continues generation as usual conditioned on both the prior concepts and the proposed plan. An overview of the approach is shown in \Cref{fig:three_level_lcm}.

\noindent Although we envision a separate (but complementary) model for planning, 
we present here an initial experiment using a simplified single-model approach
where the \lcm is trained in a multitask setting to also predict both the \emph{break} concepts and plans.
Additionally, instead of the idealized plan spanning multiple concepts (such as a paragraph), 
we use a single concept in order to capture what should come next (i.e. a \emph{plan} concept).
We call this simplified multitask approach a Large Planning Concept Model (\planlcm). 

\subsubsection*{Methodology} %

In order to evaluate this single-model approach, we perform an ablation study. As a baseline method, we train a \interleaved \lcm (cf. \Cref{sec:arch:interleaved}) without any visibility to \emph{break} or \emph{plan} concepts. We then subsequently train a \planlcm with the same number of parameters as the baseline. Both models were trained on the same data\footnote{Compared to previous experiments, we use a different data mixture more favorable for long-form generation.} for the same number of steps. %

\paragraph{Data preprocessing.} 
In order to represent \emph{break} concepts, we begin by first segmenting the data into paragraphs. 
Given that most real world datasets are absent of paragraph structure (or it is not easy to recover), 
we apply the Segment Any Text \citep{frohmann-etal-2024-segment} paragraph splitting API\footnote{\url{https://github.com/segment-any-text/wtpsplit}}. 
We additionally force each paragraph to be less than 10 sentences, and merge small (e.g. one sentence) consecutive paragraphs together.
In order to represent \emph{plan} concepts, we generate synthetic high-level topic description for each preceding segmented paragraph using an existing open-sourced \llm, namely Llama-3.1-8B-IT,
which offers a good trade-off between the generated topic quality and the generation speed. 
The system prompt used to generate these topic descriptions is listed in \Cref{sec:prompt_generation_of_topic_descriptions}. 
In total we process approximately $\textrm{320M}$
paragraphs with topic descriptions, spanning $\textrm{1.5B}$ segmented concepts (i.e. approximately $\textrm{30B}$ tokens).

\paragraph{Metrics.} We focus on coherence as our main measure of evaluation. Previous ablations (cf. \Cref{sec:archi:lcm:ablations}) used the coherence metric introduced by \citet{jwalapuram-etal-2022-rethinking}. However, we explore here \llm-as-a-judge as an alternative. Specifically, we use Llama-3.1-8B-IT in order to evaluate the coherence of the generated model outputs, which is prompted to return an overall coherence score between $\left[0, 5\right]$. 
The prompt used is listed in \Cref{sec:prompt_llm_as_a_judge_coherence}.
In order to validate this prompt, we evaluate it against a dataset of human judgements introduced by \citet{jwalapuram-etal-2022-rethinking}, and observed it reported an agreement\footnote{Krippendorff's $\alpha$ = 0.48} with human annotators which improves upon their coherence model. We therefore choose this metric for our evaluation.
To be consistent across both model results, we do not include the special \emph{break} or \emph{plan} concepts generated by the \planlcm when calculating coherence scores.

\begin{table}[ht!]
    \centering
    \begin{tabular}{cc}
    \toprule
         & Llama-3.1-8B-IT ($\uparrow$) \\
         \midrule
         \planlcm & {\bf 2.82} $\pm$ 0.62 \\
         Baseline & 2.74 $\pm$ 0.70\\
    \toprule
    \end{tabular}
    \caption{\planlcm ablation coherence score results.}
    \label{tab:hlcm_ablation_results}
\end{table}

\subsubsection*{Results} 

We provide the results of our ablation experiment in \Cref{tab:hlcm_ablation_results}. Results are reported over a held-out subset of \cosmopedia\citep{cosmopedia} following instruction fine-tuning, similar to previous ablations (cf. \Cref{sec:archi:lcm:ablations}).
We observe that the \planlcm achieves significantly\footnote{Significance observed at the 99.9\% level.} higher coherence scores (significance was measured using a paired t-test) than the baseline \interleaved \lcm. This finding suggests that the \planlcm is capable of producing significantly more coherent outputs than the \lcm as a result of the additional structure coming from predicted plan concepts, helping the \planlcm produce a more coherent narrative, which is essential for the objective of generating long-form output. 
\section{Related work}

\subsection{Sentence representations}

\paragraph{Multilingual sentence representations}

Learning effective sentence embeddings has been a well studied subject in recent years.
Significant progress has been made in this field, largely due to the capabilities of transformer-based language models that by learning contextual representations for individual tokens \citep{devlin2018bert, xlmr}, are able to effectively learn the semantics of language.
However, such models are not optimal to create sentence representations.

Following approaches built upon these initial works, and aimed at learning general sentence representations, leveraging dual encoder architectures \citep{guo2018effective, reimers2019sentence, ni2021sentence}.
These architectures encode the source and target into a common embedding space, and use a distance-based metric to create an alignment loss that approximate semantically identical sentences.
Such architectures have been extended to leverage multilingual data to create general, aligned embedding spaces across languages \citep{labse, mexma, sturua2024jinaembeddingsv3multilingualembeddingstask}.
Initial approaches leveraged the contrastive loss to align translations across languages \citep{labse, yang2019improving}, using only translation data to train.
Other architectural changes, namely using token-level objectives combined with the sentence level objectives, have proven useful to improve the quality of multilingual sentence representations based on translation data only \citep{li-etal-2023-dual, mexma}.
Recent approaches explore using data from other tasks, besides translation data, to increase the generality of the sentence representations \citep{wang2024multilinguale5textembeddings, mohr2024multitaskcontrastivelearning8192token}.
Other approaches change their embeddings per task, either with task-specific prompts \citep{wang2024multilinguale5textembeddings, su2022one, lee2024geckoversatiletextembeddings} or with task-specific parameters \citep{sturua2024jinaembeddingsv3multilingualembeddingstask}.

Another successful line of work to create general purpose, multilingual, sentence representations is to leverage the translation objective.
\textsc{LASER} \citep{Artetxe:2019:tacl_massive_ml}, and \sonar \citep{Duquenne:2023:sonar_arxiv} leverage an encoder-decoder architecture, with a fixed-size sentence representation between the encoder and the decoder, trained with a translation objective.
\sonar is initialized from the NLLB-200 model \citep{nllb2022}, and covers 200 languages, making it one of the open-source models with the widest language coverage.
\sonar also provides open-source speech encoders aligned to their sentence encoders for 73 languages \citep{seamlessv2:arxiv:2023}, aligned through a teacher-student approach.
\sonar has been used as the basis for several works \citep{seamlessv2:arxiv:2023, SeamlessM4TArXiv, chen-etal-2023-blaser}, and its speech decoders have been extended to keep the expressiveness of the original speech \citep{Duquenne:2023:sonarexp_arxiv}.

\paragraph{Joint speech/text sentence representations}
There has been a large body of research on unsupervised representation learning for monolingual \citep{wav2vec2} and multilingual speech \citep{babu2021xls}, with recently w2v-bert \citep{chung2021w2v} that combines contrastive learning and masked language modeling to learn self-supervised representations from speech. Other works explored multilingual and multimodal (speech/text) pre-training methods, including mSLAM \citep{mslam}. Finally, \citet{Duquenne:2021:nips_mine}, followed by \citet{khurana2022samu}, introduced multilingual and multimodal sentence embeddings, extending a pre-existing multilingual text sentence embedding space to the speech modality with a distillation approach. \citet{tmodules:acl,interspeech} also showed that it is possible to efficiently decode multilingual speech sentence embeddings with decoders trained on text sentence embeddings into different languages, to perform zero-shot speech translation.

\paragraph{\llm based sentence representations}

Several text representation methods have been proposed which are based on existing \llms.
\citet{wang-etal-2024-improving-text} proposed extracting text embeddings from the last token of LLMs fine-tuned with instructions on contrastive data. \citet{lee2024nv} improved text embedding capabilities of fine-tuned LLMs by removing the causal attention mask and applying extra nonlinear layers before pooling the token embeddings. 
Embeddings as a service are supported by some commercial LLM providers, for example, \textsc{Mistral-embed}.\footnote{\url{https://docs.mistral.ai/capabilities/embeddings/}}
Such embeddings proved competitive on retrieval benchmarks; however, to the best of our knowledge, their applicability to reconstructing the texts back from the embedding space has not been demonstrated.

\subsection{Multilingual \llms}

Most of the leading \llms have been trained on texts in several languages. \Cref{fig:archi:langs} summarizes the coverage of several of them. Nevertheless, the pretraining data of these \llms seems to be mainly English texts. For example, \citet{llama3.arxiv.2024} mentions that pretraining data contains significantly more English texts, requiring continued pre-training with multilingual data, out of which 34.6\% is translated reasoning data.

There are also several efforts to train \llms optimized on specific languages, e.g. \textsc{LeoLM} for German,\footnote{\url{https://laion.ai/blog/leo-lm/}} \textsc{Fuano}
for Italian \citep{fuano:2024:arxiv}, \textsc{ALLaM} for Arabic \citep{allam:2024:arxiv}, and several models for Chinese: \textsc{ErniBot},\footnote{\url{http://research.baidu.com/Blog/index-view?id=183}} \textsc{Tongyi Qianwen},\footnote{\url{https://www.alibabacloud.com/en/solutions/generative-ai}} or \textsc{ChatGLM} \citep{chatglm:2024:arxiv}. Some adaptations of LLMs to a massive number of languages also exist. LOLA \citep{srivastava2024lola} is a recent mixture-of-experts LLM supporting 160 languages, MALA-500 \citep{ji2024emma} adapts LLaMA2 to 546 languages. However, such models typically face a trade-off between language coverage and other capabilities. For example, the Aya model \citep{ustun2024aya}, following instructions in 101 languages, was superseded by Aya-23 \citep{aryabumi2024aya} that exchanged some breadth for depth, focusing on 23 languages only. The LCM architecture, combining a language-agnostic model for knowledge and reasoning with potentially language-specialized encoders and decoders, is expected to exhibit this trade-off to a lesser extent.

\subsection{Alternative \llm architectures}

Predicting the next state in the embedding space is a core idea of the Joint Embedding Predictive Architecture (\jepa) proposed by \citet{jepa:openreview:2022}. This idea has been implemented for images (\textsc{I-JEPA} by \citet{ijepa:arxiv:2023}) and video (\textsc{V-JEPA} by \citet{vjepa:arxiv:2023}) as a self-supervised approach to learning representations. For language, equivalent models have not yet been explored.

\paragraph{Sentence embeddings for language modeling.} For text completion, \citet{storyline-llm:arxiv:2020} proposed a sentence-level language model operating by choosing the next sentence from a finite set of candidates. Their model demonstrated success in selecting appropriate continuations for short stories, but it has not been scaled to longer inputs or to fully generative outputs. \citet{golestani2021using} studied a similar problem in the even more restrictive sentence ordered setting, but with a more thorough study of architectural choices. The INSET architecture \citep{inset:acl:2020} solves the sentence infilling task by combining a denoising autoencoder that encodes sentences into fixed-size vectors and decodes them back and a bidirectional transformer that predicts the embedding of a missing sentence. 

\citet{newsum:acl:2021} and \cite{cornille2024learning} used predicted next sentence embeddings in a fully generative setting, for summarization and generic language modeling, respectively. However, their architectures considered sentence-level connections only as an addition to the token-level connections across sentences, not as their replacement. 

In a recent work of \citet{an2024sentencevae}, the SentenceVAE architecture performs language modeling on the sentence level using a sentence encoder to prepare the inputs and a sentence decoder to produce the outputs. However, its input and output embedding spaces are not tied, so the inference is only possible by decoding each predicted sentence into text and then re-encoding it for adding it to the context.

\paragraph{Language modeling with diffusion.} A series of more recent works tried adapting diffusion modeling, originally developed for continuous data, to the discrete text domain. The PLANNER architecture \citep{planner:neurips:2023} consists of a variational autoencoder for paragraphs and a diffusion model trained to predict latent autoencoder representations conditional on the textual context or on the class label. \citet{diffguidlm:arxiv:2024} augmented a decoder-only language model with an encoded semantic proposal of the continuation text, with an easily guidable diffusion model predicting the embedding of the next proposal. A TEncDM model \citep{tencdm:arxiv:2024} performs diffusion in the space of contextual token embeddings which are then decoded non-autoregressively.

Some applications of diffusion to sequence modeling have targeted the planning capabilities of the sequence models. Semformer \citep{semformer:arxiv:2024} proposed training transformers language models to plan several steps ahead by including special planning tokens, the representations of which are trained to be informative about the future tokens. \citet{discdiff-reason:arxiv:2024} applied discrete diffusion to language models as an alternative to autoregressive generation, more suitable for tasks that require multi-step planning. \citet{diffplan:arxiv:2024} give an overview of applications of diffusion for planning tasks, but most of them are not concerned with the language domain.

Overall, while many of the previous works used hidden representations for language modeling or related tasks, all of them either relied on token-level inputs or outputs, or were not intented for generating texts of arbitrary length. The \lcm seems to be the first fully generative language model implemented fully in a highly semantic, reconstructable sentence representation space.

\section{Limitations}
\label{sec:limits}

In this section we discuss the possible limitations of the presented Large Concept Modeling approach.

\paragraph{Choice of the embedding space.}
 The choice and design of the embedding space plays a crucial role in the \lcm modeling approach. 

\begin{itemize}
    \item The \sonar embedding space was chosen for its good multilingual and multimodal representations, as well as the availability of a massively multilingual decoder, which achieves excellent results in both translation and auto-encoding. However, the \sonar model was trained on very specific training data, namely bitext machine translation data containing rather short sentences. This has several consequences:
    \begin{enumerate}
        \item \sonar is trained to sustain a local geometry (sentences with very similar meanings are geometrically close) with no special guarantees for sentences that are only loosely related.
        Yet, predicting next sentences distribution requires the space to operate well globally. 
        \item \sonar auto-encodes surprisingly well texts containing links, references, or merely numbers or code data.
        Yet, such texts tend to be fragile, highlighting a distribution mismatch between the \sonar training data and commonly used \llm pre-training text corpora.
        Therefore, the accurate prediction of the sentences containing such a content (non-negligible in \lcm pre-training data) will be hard for any \lcm \sonar based model. For instance, the factuality of fragile generated sentences may easily be compromised. 
    \end{enumerate}

\item   Using a frozen encoder represents some interesting trade-offs. 
Any frozen encoder which is learned in a different data context, and with no a-priori strong connection to \lcm modeling, may be suboptimal compared to encoders that are learned in an end-to-end fashion (with the loss coming from the decoder). 
 At the same time, learning an encoder within end-to-end training can be challenging and the resulting space is not guaranteed to result in good semantic representations shared across languages and modalities.

 Training the concept representation and the \lcm end-to-end would also be less data and compute efficient since all modeling data should be multilingual and -modal, bearing the risk of modality competition.
\end{itemize}

\newpage
\paragraph{Concept granularity}
\begin{itemize}
    \item In this work, the definition of concepts is interpreted at sentence level. However, the manifold of possible next sentences is very wide, attributing a proper probability to each of such sentences is much harder (even with a modeling within the latent space) that to the discrete set of tokens.
    \item In NLP, we encounter sentences of variable length. Combinatorial complexity of possible next sentences grows exponentially with the maximum character length. 
    The choice of granularity for \lcm is not trivial as long sentences (>120 characters) could reasonably be considered as several concepts.
    However, any finer splitting of such sentences does not necessary separate well these concepts. This shows the limitation of a fixed size embedding representation for one sentence. Text splitting (such as sentence splitting) or one-to-many mapping of a sentence into several embeddings is a major future direction of research.
    \vspace{-0.5mm}
    \item Each document in a training corpus typically contains a sequence of unique sentences or a little number of repetitions. This data sparsity effect manifests as well at large corpora level: the large majority of sentences are merely unique. In principle, this issue can be addressed with higher-level semantic embedding representations. These higher-level representations come with trade-off between requirement of lossless data encoding (think of named-entities or numbers, critical in many language modeling tasks) and  good level of abstraction to enable reasoning capabilities. Compared to a monolingual auto-encoder which would simply compress input sentences, \sonar offers semantic representations with good auto-encoding quality but still certainly sacrificing generalization capacities.
    \vspace{-0.5mm}
    \item This generalization issue can be partially mitigated by splitting or encoding input text as new conceptual units which are more commonly shared across source documents. This is in the spirit of stemming or lemmatization techniques studied in NLP for words. 
    That being said, building such conceptual units that are also language and modality agnostic is a challenging task. Such shared multilingual and multimodal conceptual units are also key for generalization across languages and across modalities. To maximize cross-lingual and cross-modal transfers, \LCMs should be exposed to a richer variety of multilingual and multi-modal data.

\end{itemize}

\vspace*{-4mm}
\paragraph{Continuous versus discrete} 
\begin{itemize}
    \item  Diffusion modeling has proven to be very efficient in generative modeling of continuous data like images or speech. As previously stated, sentences in the \sonar space, despite being represented as continuous vectors, remain discrete combinatorial objects. This makes diffusion modeling struggle on the text modality (either at word or sentence embedding level).
    \vspace{-0.5mm}
    \item The contrastive nature of cross-entropy loss based on softmax outputs which is used for next token prediction plays a critical role for many downstream task where higher accuracy is required (e.g. MCQ tasks, code or math generation).
    On the opposite, continuous diffusion modeling does not allow to integrate such a contrastive objective.
    \vspace{-0.5mm}
    \item The \qlcm could be a way to address the discrete nature of text while modeling on coarse-to-fine semantic units shared across languages and modalities. The limited performance of the \qlcm approaches presented in this paper may be explained by the fact that \sonar space was not trained to be efficiently quantizable, yielding a significant number of codebooks and a large amount of units per codebook. Therefore, the current \sonar quantization suffers from the exponentially increasing number of RVQ units combinations which does not solve the data sparsity/uniqueness issue discussed earlier. 
    This indicates once again the importance of developing a new representation space, either continuous or discrete, for the \LCM.  

\end{itemize}

\section{Acknowledgments}
We would like to thank
Robbie Adkins,
Can Balioglu,
Joy Chen,
Pascale Fung,
Jason Holland,
Amita Kamath,
Justine Kao,
Sagar Miglani,
Alice Rakotoarison,
Abhilasha Sancheti,
Arjang Talattof,
Ellen Tan,
Carleigh Wood,
Shireen Yates,
Bokai Yu and
Luke Zettlemoyer
for comments and suggestions on this work, as well helping us to improve this paper.

\section{Conclusion and Future Work}
\label{sect:concl}
Current best practice for large scale language modeling is to operate at the token level, i.e. to learn to predict the next tokens given a sequence of preceding tokens. There is a large body of research on improvements of \llms, but most works concentrate on incremental changes and do not question the main underlying architecture.
In this paper, we have proposed a new architecture, named a \LCM (\lcm), which substantially differs from current \llms in two aspects:
1)~all modeling is performed in a high-dimensional embedding space instead of on a discrete token representation;
and 2)~modeling is not instantiated in a particular language or modality, but at a higher semantic and abstract level. We have named the general form of this representation a~\textit{``concept''}.

In this paper, to verify the feasibility of the high-level idea, we have assumed that a concept corresponds to a sentence in the text domain, or an equivalent speech segment, and that the embeddings are obtained by the freely available \sonar sentence encoder \citep{Duquenne:2023:sonar_arxiv}. With respect to the specific architecture of the \lcm, we have first shown that directly minimizing the MSE loss in the embedding space does not yield good results. We then explored several architectures based on a diffusion process: the \interleaved and \twotower \lcm, as well as a \qlcm which uses quantization of SONAR representations and then modeling on these discrete units.
These ablation experiments were performed with models with 1.6B parameters and focused on the generative task of continuing a sequence of sentences.
We have then scaled our models to a size of 7B parameters and instruction-finetuned them on several summarization and summary expansion tasks. We provide a detailed comparison to other public models of the same size, namely \gemma, \mistral and \llama.

By design, a \lcm exhibits strong zero-shot generalization performance. In this paper, we trained models on English texts only, and applied them to text in other languages, without any additional training data, neither aligned nor unlabeled.
The \lcm outperforms \llamaIT on English and on the average over foreign languages officially supported by the \llm.
The \lcm itself could also be trained on multilingual- and model data to acquire knowledge from these sources. We will explore this in future versions of the \lcm.
In short, all languages and modalities are first class citizens and handled equally at all stages of a \lcm.

We have observed that next sentence prediction is substantially more challenging than next token prediction.
First, given that we operate in an embedding space and at a higher semantic level, the number of possible sentences is virtually unlimited, while token vocabularies are usually in the range of 100k. Second, even given a long context, there is unavoidably more ambiguity in choosing the next sentence than the next token. And third, the usual softmax output layer over the fixed size token vocabulary provides a normalized probability distribution over all possible token continuations. Theoretically, a diffusion process should be able to learn a probability distribution over an output embedding space, but our current experimental evidence indicates that more research is needed to take full advantage of the properties of \LCMs. As an example, the ability to sample multiple embeddings and associate a score would enable beam search to find the best sequence of sentences.
Finally, small modeling errors could yield predictions in the embedding space which do not correspond to valid sentences, i.e. that cannot be decoded into a syntactically and semantically correct sentence. We will work on alternative concept embeddings to \sonar which would be better suited to the next sentence prediction task, and would improve modeling approaches in that concept embedding space.

We see the models and results discussed in this paper as a step towards increasing scientific diversity and a move away from current best practice in large scale language modeling.
We acknowledge that there is still a long path to reach the performance of current flagship \llms. This will require of course further improving the core architecture, but also careful data selection and curation, extensive ablations, optimized and diverse instruction fine-tuning, and finally, scaling to models with more than 70B parameters.

We open-source the full training code of all our \lcm variants, together with a set of supporting scripts,\footnote{\meresgithub} to make it easy for other teams to train \lcm models.
By these means, we hope to foster research on alternative \llms and contribute to advance the field of machine intelligence.
\bibliography{bib-meres,bibliography}%
\newpage

\appendix
\addcontentsline{toc}{section}{Appendices}
\addtocontents{toc}{\protect\setcounter{tocdepth}{0}}

\section[Data]{Technical consideration for data preparation}
\label{app:data:technic}

Since our modeling approach uses a fixed encoder and a
fixed document segmentation method, 
we decided to use pre-computed \sonar embeddings instead 
of producing them on-the-fly for each training run. This allows for faster iteration on the same data mix, trading expensive GPU compute against storage capacity.

As we are storing sequences of \sonar embedding, which are fixed size tensors of 1024 floats, the storage requirements become more demanding than storing the raw text. For one terra bytes of raw text data we need to store between fifteen and twenty terra bytes of encoded data.
Overall, this trade-off in space vs compute reduces the GPU memory occupation and the compute load and lets use iterate faster.
Typically, with on single GPU we can produce around 
300-400 \sonar sentence embeddings per second whereas 
by loading precomputed data (potentially from remote storage) 
we can load over 20 thousand embeddings per second per GPU (with around 15 CPU per GPU).

We store sequences of \sonar embeddings with 16 bits precision (FP16) in parquet datasets.
Embeddings remain aligned with the segmented texts and the
parquet binary format and library ecosystem is well suited for storing and loading efficiently such complex data structures. Parquet also lets us store extra data (such as quality metrics for each sentences) and enables non-trivial last mile data filtering and transformation.

For training the \lcm, we processed around four billion documents, generating 310 billion sentences with an average of 27 tokens per sentences for 88 characters length on average; totaling a bit more than 889 terra-bytes of raw text.\section{Open Sourced Code}

In the spirit of reproducibility, we release under an open source license the training, evaluation and data processing code for the \lcm. This is available at \meresgithub.

The training code is based on the \fairseq framework \citep{balioglu2023fairseq2} that allowed us to build and iterate over the different model architectures discussed above. While \fairseq shares the same name as the popular fairseq toolchain, its API architecture is different. It is not a monolithic toolchain but a set of modules that can be composed, this allows us to build different architectures side by side and easily share training components.

We also release our evaluation framework so the evaluation tasks reported in \ref{sec:task} and comparison between the \lcm and other models can easily be reproduced. The evaluation framework provides a clear abstraction between \emph{predictors}, \emph{tasks} and \emph{data loading}, which again, makes it modular and lets us describe a set of tasks to be evaluated. The evaluation framework can be run locally or distributed over a SLURM cluster to run evaluations at scale.

Finally, we release an updated version of stopes\footnote{\stopesgithub} to simplify large scale data pre-processing on a SLURM cluster. This was used to run the sentence segmentation and \sonar encoding described in section \ref{sec:data}. The stopes data processing framework deals with scheduling and monitoring large number of jobs on SLURM or to run everything locally for small scale jobs. It provides an API compatible with \texttt{ray.data}\footnote{\url{https://docs.ray.io/}} that makes it easy to process large datasets in blocks and apply transform function over it. This makes our code reusable outside of a SLURM cluster as it can also be used with a ray.io cluster.\section{System prompt: Generation of Topic Descriptions}
\label{sec:prompt_generation_of_topic_descriptions}

You are a topic description generator.
Your job is to read an extract of text and then generate a topic description.
The extract may be well formed or not.
The topic description you will write will be at most one sentence in length, and use as few words as possible.
However, it can not be generic and it can not contain any profanity.

Here is an example of an extract, an ideal topic description, and some examples of bad topic descriptions:

Example extract: “One day, one neighborhood of the city was completely devasted. Glass windows were shattered, shops turned upside down, and many civilian killed. Superman instantly recognized the signature of one of his old enemies, Voltar, who he had barely beaten in the past. This was a message to him: "I challenge you! Come find me!""

An example of a good topic description: An old enemy of Superman's, Voltar, appeared and challenged him.\\
An example of a bad topic description: Superman\\
An example of a bad topic description: Voltar\\
An example of a bad topic description:

\section{User prompt: LLM As a Judge - Coherence}
\label{sec:prompt_llm_as_a_judge_coherence}

Below is a text extract. Your task is to analyze the extract and assign a coherence score between 0 and 5 inclusive, where:

0: The text is completely incoherent and lacks any logical connection.\\
1: The text has some minor connections, but overall it is disjointed and hard to follow.\\
2: The text has some coherence, but it is still difficult to understand due to unclear relationships between ideas.\\
3: The text is moderately coherent, with some clear connections between ideas, but may lack depth or clarity.\\
4: The text is highly coherent, with clear and logical connections between ideas, making it easy to follow.\\
5: The text is extremely coherent, with a clear and concise structure, making it effortless to understand.

You will provide a score ONLY. Do NOT also provide an explanation.

The extract: <extract>

After examining the extract, the coherence score between 0 and 5 inclusive is:

\restoregeometry
\end{document}